\documentclass[10pt, leqno]{article}
\usepackage[english]{babel}

\usepackage[utf8]{inputenc}
\usepackage[T1]{fontenc}
\usepackage{dsfont}
\usepackage{hyperref}       % hyperlinks
\usepackage{url}            % simple URL typesetting
\usepackage{booktabs}       % professional-quality tables
\usepackage{amsfonts}       % blackboard math symbols
\usepackage{nicefrac}       % compact symbols for 1/2, etc.
\usepackage{amsmath, amssymb}
\usepackage[a4paper,hscale=0.75,vscale=0.8,centering]{geometry}
\usepackage{setspace}
\usepackage{authblk}
\usepackage{amsfonts}
\usepackage{bbm}
\usepackage{enumitem}
\usepackage{adjustbox}
\usepackage[font=small]{caption}
\usepackage{dsfont}
\usepackage{graphicx}
\usepackage{float}
\usepackage{xcolor}
\usepackage{wrapfig}
\usepackage{tikz}
\usepackage{pst-node}
\usepackage{booktabs} 
\usepackage{multirow}
\usepackage{makecell} 
\usepackage{algorithm}
\usepackage{algorithmicx}%
\usepackage{algpseudocode}%
\usepackage{listings}
\usepackage{fancyvrb}
\usetikzlibrary{fit,positioning,bayesnet}
\usetikzlibrary{automata,arrows}
\usetikzlibrary{backgrounds}
\usetikzlibrary{arrows.meta}

\definecolor{darkred}{rgb}{.7,0,0}
\definecolor{darkgreen}{rgb}{0,0.7,0}

\def\f0{{\mathbf 0}}

\usepackage{subfigure}

\usepackage{natbib}
\usepackage{hyperref}
\definecolor{bred}{rgb}{0.8,0,0}
\hypersetup{colorlinks,linkcolor={blue},citecolor={black},urlcolor={blue}}
\setcitestyle{square,numbers,comma}
\usepackage{cleveref}

\def \y {\mathbf y}

\def \0{\mathbf 0}

\def\b0{\mathbf 0}

\def \y{\mathbf{y}}

\def \bfy {\mathbf{y}}
\usepackage{xcolor}
\usepackage{afterpage}
\usepackage{appendix}
\usepackage{bbm}
\usepackage{tikz}
\usetikzlibrary{fit,positioning,bayesnet}
\usetikzlibrary{automata,arrows}
\usetikzlibrary{backgrounds}
\usepackage{amsthm}

\def\genbox#1#2#3#4#5#6{% #1=0/1, #2=color, #3=shape, #4=raise, #5=width, #6=width/2
    \leavevmode\raise#4bp\hbox to#5bp{\vrule height#5bp depth0bp width0bp
    \pdfliteral{q .5 w \csname #2COLOR\endcsname\space RG
                       \csname #3PDF\endcsname{#5}{#6} S Q
             \ifx1#1 q \csname #2COLOR\endcsname\space rg 
                       \csname #3PDF\endcsname{#5}{#6} f Q\fi}\hss}}
\title{{\scshape{Hierarchical Inference and Closure Learning via Adaptive Surrogates for ODEs and PDEs}}}

\newcommand{\cblue}{\textcolor{blue}}

\author[$\cblue{\ddagger}$]{Pengyu Zhang \footnote{Corresponding authors \{pz281, av537\}@cam.ac.uk.}}
\author[$\cblue{\ddagger}, $]{Arnaud Vadeboncoeur$^*$}
\author[$\cblue{\ddagger}$]{Alex Glyn-Davies}
\author[$\cblue{\ddagger, \S}$]{Mark Girolami}

\affil[$\cblue{\ddagger}$]{\small \textit{Department of Engineering, University of Cambrdige, 7a JJ Thomson Ave, Cambridge, CB3 0FA, UK}}
\affil[$\cblue{\S}$]{\textit{The Alan Turing Institute, 96 Euston Rd., London, NW1 2DB, UK}}
\date{}

\begin{document}

\maketitle

% REQUIRED
\begin{abstract}
Inverse problems are the task of calibrating models to match data. They play a pivotal role in diverse engineering applications by allowing practitioners to align models with reality. 
In many applications, engineers and scientists do not have a complete picture of i) the  detailed properties of a system (such as material properties, geometry, initial conditions, etc.); ii) the complete laws describing all dynamics at play (such as friction laws, complicated damping phenomena, and general nonlinear interactions). 
In this paper, we develop a principled methodology for leveraging data from collections of distinct yet related physical systems to jointly estimate the individual model parameters of each system, and learn the shared unknown dynamics in the form of an ML-based closure model.
%
% This paper develops a principled framework for solving challenging inverse problems involving unknown parameters and nonlinear closures where data is collected for a collection of related yet distinct physical systems. 
%
To robustly infer the unknown parameters for each system, we employ a hierarchical Bayesian framework, which allows for the joint inference of multiple systems and their population-level statistics.
To learn the closures, we use a maximum marginal likelihood estimate of a neural network embeded within the ODE/PDE formulation of the problem.
To realize this framework we utilize the ensemble Metropolis-Adjusted Langevin Algorithm (MALA) for stable and efficient sampling. 
% Concurrently, nonlinear closure terms are modeled deterministically via Multilayer Perceptrons (MLPs). 
%
To mitigate the computational bottleneck of repetitive forward evaluations in solving inverse problems, we introduce a bilevel optimization strategy to simultaneously train a surrogate forward model alongside the inference.
Within this framework, we evaluate and compare distinct surrogate architectures, specifically Fourier Neural Operators (FNO) and parametric Physics-Informed Neural Network (PINNs). 
\end{abstract}

% REQUIRED
% \begin{keywords}

% \end{keywords}

\section{Introduction}\label{sec1}

Modeling and understanding physical systems governed by ordinary differential equations (ODEs) and partial differential equations (PDEs) form the cornerstone of modern scientific and engineering analysis. Such equations describe the temporal and spatial evolution of physical processes in a variety of fields, including fluid and solid mechanics \citep{timoshenko_1970,batchelor_2000}, heat transfer \citep{carslaw_1959}, electromagnetism \citep{jackson_1999}, chemical kinetics \citep{murray_2002}, and atmospheric and oceanic dynamics \citep{vallis_2017}. While these equations are often derived from first principles, in many complex real-world scenarios, they are incomplete: the governing equations may contain unknown parameters, or there are missing/unknown nonlinear closures. Accurately determining both the unknown parameters and closures from observational data, as an inverse problem, is crucial for model calibration and enabling accurate predictions of physical dynamics for application. Despite the ill-posed nature of inverse problems, the increasing availability of observational data and the development of modern machine learning techniques to identify complex patterns have improved the ability to solve inverse problems for discovering unknown parameters or physical terms directly from data.

In many practical systems, the governing equations are mostly known and only a portion of the dynamics is poorly understood.
This is in contrast to system identification or equation discovery frameworks such as the sparse identification of nonlinear dynamics (SINDy) approach \citep{brunton2016discovering,rudy2017data,messenger2021weak,champneys2025bindy}, which aim to recover the entire governing equation from data, our method focuses on cases where part of the physical model is already known. 
Common examples of these unmodeled components include friction in bearing systems, complicated damping laws in mechanisms, turbulence models in fluid dynamics, or nonlinear heat dissipation due to convection \citep{duraisamy2019turbulence, safari2023data,sirignano2023deep}. In a similar context, Rogers \textit{et al.} \cite{rogers2020application, rogers2022latent} address unknown nonlinear dynamics in mechanical systems by modeling the missing terms as latent restoring forces using Gaussian Processes (GP) \cite{rasmussen2003gaussian}, where same method is used in identifying nonlinear convection effects in heat transfer \cite{kouw2024bayesian}. Following a similar philosophy, rather than discovering the entire equation from scratch, our approach leverages the known physics and infers only missing components.

To improve the quality of inference we adopt a population-based formulation in which observational data are gathered from multiple systems belonging to the same physical family. These systems, indexed by $k=1,...,K$, have distinct parameters $\boldsymbol{\theta}^{(k)}$, yet share a common nonlinear closure function $f(\cdot)$. The underlying physical model can be expressed through a parameter-to-solution operator $F^\dagger : \Theta \rightarrow U $ which maps the pair $(\boldsymbol{\theta}^{(k)},f)$ to the corresponding $k^{\text{th}}$-system state,
\begin{equation}
    \label{eq:general_system}
    F^\dagger(\boldsymbol{\theta}^{(k)},f) = \mathbf{u}^{(k)}.
\end{equation}
This general formulation encompasses a broad class of physical systems that our proposed framework can handle, including ODEs and PDEs. For each system $k$, sparse and noisy observations $\mathbf{y}^{(k)}$ from system $k$ are collected via
\begin{equation}
    \label{eq:observation_operation}
    \mathbf{y}^{(k)}=G^{(k)}(\boldsymbol{\theta}^{(k)},f)+\boldsymbol{\xi}_\eta ^{(k)}, \quad G^{(k)} = g^{(k)} \circ F^\dagger,
\end{equation}
and $g:U \rightarrow \mathbb{R}^{d_y}$ is a solution-to-data map, typically a linear map which can be represented by an observation matrix $\mathbf{H}^{(k)}$. The noise term is assumed independent across systems, with $\boldsymbol{\xi}_\eta ^{(k)} \sim \mathcal{N}(0, \sigma^{(k)}\mathbf{I})$ i.i.d.. The objective of the proposed method is to simultaneously infer unknown system-specific parameters $\boldsymbol{\theta}^{(1:K)}$ and to learn the shared closure model $f$, given observational data $\bfy^{(1:K)}$. 

% Numerous optimization-based approaches have been developed to infer unknown parameters in ODE and PDEs from observational data, including  \citep{aarset2023learning,raissi2019physics,arridge2019solving}. Despite their success, they are typically deterministic, providing only point estimates of parameters without quantifying their uncertainty. 
A wide range of deterministic methods have been developed for parameter inference in differential equations \citep{raissi2019physics,arridge2019solving,aarset2023learning}. While effective in recovering optimal parameters, these frameworks generally focus on point estimation and lack inherent mechanisms for uncertainty quantification (UQ). However, UQ is essential in inverse problems, due to their inherent ill-posedness, it is necessary to characterize the distribution of possible parameters rather than relying on a single estimate. To incorporate UQ, the Bayesian framework has been widely adopted in inferring unknown parameters in physical systems \citep{stuart2010inverse,cotter2010approximation,bui2013computational,petra2014computational,bui2014solving}. In this framework, the inverse problem is formulated probabilistically: given observational data with their uncertainties, a forward model, and a prior distribution, the goal is to recover the full posterior distribution of the unknown parameters. Building on this idea, we utilize a hierarchical Bayesian model \citep{gelman1995bayesian,chib2018hierarchical} that jointly infers parameters of multiple related systems. This hierarchical formulation captures both system-specific variability and shared population-level structure, improving inference stability and enabling knowledge transfer across systems.

Whilst a Bayesian approach to inference of the system parameters $\boldsymbol{\theta}^{(1:K)}$ is feasible since they are low-dimensional, the unmodeled nonlinear closure cannot be easily expressed using a small number of parameters, making a Bayesian appoach computationally prohibitive. To represent the nonlinearity, we introduce a neural network approximation, training a Multi-Layer Perceptron (MLP) with parameters $\alpha$ such that $f^{\alpha}\approx f$, which serves as a flexible neural representation of the nonlinear closure.

In summary, our proposed method utilizes a hierarchical Bayesian formulation coupled with ensemble Metropolis-adjusted Langevin algorithms (MALA) \citep{roberts1996exponential} for probabilistic sampling of system parameters $\boldsymbol{\theta}^{(1:K)}$, and a deterministic neural network to learn the unmodeled nonlinear closure $f$. We alternate between probabilistic sampling and closure model updates in an iterative scheme. This hybrid approach achieves a balance between statistical interpretability and computational efficiency. The physical parameters $\boldsymbol{\theta}^{(1:K)}$ are low-dimensional, making them well-suited for probabilistic inference with UQ. In contrast, the nonlinear closure is high-dimensional and complex, where deterministic training allows for efficient learning, avoiding the prohibitive cost of sampling in function space.

In addition, solving inverse problems is often computationally inefficient due to the necessity of repeatedly evaluating the forward model, which typically relies on expensive numerical solvers. This process becomes particularly prohibitive for nonlinear and time-dependent systems, especially when employing gradient-based samplers like MALA that require differentiating through the numerical solver. To alleviate this bottleneck, recent advances in surrogate modeling have been widely adopted to accelerate the simulation of physical systems. Prominent approaches include Gaussian Processes (GPs) \citep{rasmussen2003gaussian,stuart2018posterior,bai2024gaussian}, as well as deep learning-based architectures such as Deep Neural Networks (DNNs) \citep{deveney2019deep,yan2019adaptive}, Physics-Informed Neural Networks (PINNs) \cite{raissi2019physics,li2024physics}, Fourier Neural Operators (FNO) \citep{li2021fourier,li2024physics}, DeepONets \citep{lu2021learning,wang2021learning}, and Transformers \cite{vaswani2017attention,holzschuh2025pde}. Neural surrogates can significantly accelerate forward simulations. Thus, we further propose a surrogate-accelerated inference framework, in which a differentiable neural surrogate is trained jointly with the inverse problem to approximate the forward operator. The surrogate enables fast and differentiable approximations of the forward solver, thereby reducing computational cost. Our method addresses both forward surrogate training and inverse inference in a bilevel optimization framework \citep{sinha2017review}, where the lower level trains the forward surrogate model and the upper level deals with the inverse problem using the surrogate in place of the expensive numerical solver. 

\subsection{Related Work}

In this section, we summarize prior work on surrogate-accelerated inverse problems. Surrogates have been incorporated into inverse problems in two major ways: deterministic and probabilitistic surrogate-based inversion. PINNs and neural operators have been embedded into deterministic inverse formulations to replace the expensive PDE solver \citep{raissi2019physics,behroozi2025sensitivity,cho2024physics,lu2022multifidelity}. Similar to our proposed method, Zhang \textit{et al.} \cite{zhang2026bilo} develop a bilevel optimization framework, BiLO, that jointly learns the forward operator and solves the associated inverse problem, though their formulation remains entirely deterministic. In Bayesian formulations of inverse problems, PINNs and neural operators have been employed as differentiable forward surrogates to facilitate Monte Carlo sampling. For example, Bayesian PINNs (B-PINNs) treat both network weights and physical parameters as random variables, enabling posterior inference via Hamiltonian Monte Carlo \citep{neal2011mcmc} or stochastic-gradient Markov Chain Monte Carlo (MCMC) methods \citep{yang2021b, lin2022multi, li2023surrogate}. Zhang \textit{et al.} \cite{zhang2026bayesbilo} have also extended their BiLO framework for UQ in inverse problems using MCMC. Variational inference (VI) has likewise been integrated into these architectures, Yang \textit{et al.} \cite{yang2021b} also investigates VI for PINNs, while Raj \textit{et al.} \cite{raj2025deep} explores VI within DeepONet frameworks. More broadly, Glyn-Davies \textit{et al.} \citep{glyn2025primer} provide a comprehensive overview of how variational methods can be applied to both forward and inverse problems. Closely related to our approach, Akyildiz \textit{et al.} and Vadeboncoeur \textit{et al.} \cite{akyildiz2025efficient, vadeboncoeur2025population} introduced a bilevel optimization framework designed for the joint inference of prior distributions and the training of surrogate models.

Different from existing methods that typically focus on a single-system inverse problem and assume that the form of the governing ODE or PDE is fully known, our proposed method performs hierarchical Bayesian inference across multiple related systems with large collections of observations and also learns an unknown nonlinear closure model in the equation.

\subsection{Contributions}

The main contributions of this work can be summarized as follows:

\begin{enumerate}[label=\textbf{C\arabic*.}, ref=C\arabic*, leftmargin=*]
    \item \textbf{Joint probabilistic parameter estimation and deterministic closure learning.} \label{con:c1} \\
    We propose a hybrid framework that simultaneously infers unknown physical parameters and learns nonlinear closures in partially known ODEs and PDEs. Physical parameters are inferred probabilistically in a hierarchical Bayesian formulation, while the closure model is learned deterministically using an MLP.

    \item \textbf{Iterative scheme for hierarchical parameter inference and closure learning.} \label{con:c2}\\
    We implement a training strategy that alternates between posterior sampling and closure model updates. By leveraging ensemble MALA, we achieve robust sampling across multiple physical systems. These samples also serve to effectively approximate the gradients required for optimizing the closure parameters.

    \item \textbf{Surrogate-accelerated Bayesian inversion via bilevel optimization.} \label{con:c3} \\
    We incorporate a bilevel optimization scheme in which a differentiable neural forward surrogate is trained jointly with the inverse problem. This surrogate-accelerated framework greatly reduces the computational cost of Bayesian sampling by replacing expensive numerical solvers with efficient surrogate alternatives.

    \item \textbf{Validation on representative ODEs and PDEs.} \label{con:c4}\\
    We validate the proposed framework on three distinct physical problems: a nonlinear mass-spring-damper system (ODE) with an unknown damping closure, a nonlinear Darcy flow system (PDE) with an unknown diffusivity closure, and a generalized Burgers' equation (PDE).
\end{enumerate}

The rest of the paper is structured as follows. Section \ref{sec3} summarizes backgrounds underlying our approach, with a focus on Hierarchical Bayes and ensemble MALA. Section \ref{sec4} presents the proposed methodology in detail. Section \ref{sec5} demonstrates the performance of the proposed approach on three representative ODE and PDE systems. Finally, Section \ref{sec6} concludes the paper and discusses potential future directions.

\section{Background}\label{sec3}

This section presents two theoretical foundations underpinning our framework: Hierarchical Bayes and ensemble MALA. Specifically, the hierarchical Bayesian formulation provides the statistical structure to jointly model all systems. Ensemble MALA is employed as the posterior sampling method over the hierarchical model.

\subsection{Hierarchical Bayes}\label{subsec1}

Many applications involve multiple systems that are intrinsically linked by the structure of the problem. Consequently, rather than treating these systems as strictly independent, a hierarchical joint probability model should be employed. This allows the systems to share information through a common population distribution, effectively capturing their underlying dependence \citep{gelman1995bayesian}. A classical example is that the recovery probabilities of patients in the same hospital are statistically related rather than independent due to the fact that patients are subject to similar environmental conditions. Modeling each patient in isolation can therefore produce unstable estimates, especially when individual data are sparse. To address this, a hierarchical prior introduces population-level hyperparameters that characterize the group-wide behavior and allow for information to be shared across individuals.

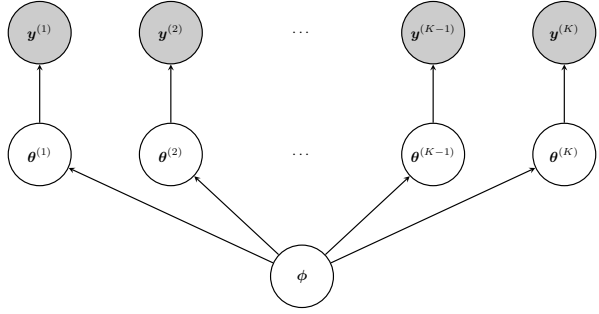
\begin{figure}[t!]
\centering
\begin{adjustbox}{max width=0.5\textwidth} % ← this actually scales the figure
\begin{tikzpicture}[
    node/.style={circle, draw, minimum size=1.5cm}, 
    obs/.style={circle, draw, fill=black!20, minimum size=1.5cm},
    >=stealth, thick
]

\matrix[row sep=1.4cm, column sep=1.6cm] {
    \node[obs] (D1) {$\boldsymbol{y}^{(1)}$}; &
    \node[obs] (D2) {$\boldsymbol{y}^{(2)}$}; &
    \node (dots1) {$\cdots$}; &
    \node[obs] (Dn1) {$\boldsymbol{y}^{(K-1)}$}; &
    \node[obs] (Dn) {$\boldsymbol{y}^{(K)}$}; \\

    \node[node] (t1) {$\boldsymbol{\theta}^{(1)}$}; &
    \node[node] (t2) {$\boldsymbol{\theta}^{(2)}$}; &
    \node (dots2) {$\cdots$}; &
    \node[node] (tn1) {$\boldsymbol{\theta}^{(K-1)}$}; &
    \node[node] (tn) {$\boldsymbol{\theta}^{(K)}$}; \\

    & & \node[node] (phi) {$\boldsymbol{\phi}$}; & & \\
};

\draw[->] (t1)--(D1);
\draw[->] (t2)--(D2);
\draw[->] (tn1)--(Dn1);
\draw[->] (tn)--(Dn);

\draw[->] (phi)--(t1);
\draw[->] (phi)--(t2);
\draw[->] (phi)--(tn1);
\draw[->] (phi)--(tn);

\end{tikzpicture}
\end{adjustbox}

\caption{Hierarchical Bayesian model illustrating how a global hyperparameter $\boldsymbol{\phi}$ governs a collection of task-specific parameters $\boldsymbol{\theta}^{(1:K)}$, each of which generates an associated dataset $\boldsymbol{y}^{(k)}$.  }
\label{fig:hier_bayes}
\end{figure}

A similar rationale applies in physical systems considered in our work. We observe multiple instances of a physical system family, such as several mass-damper systems with different masses, each governed by the same physical law but possessing distinct unknown parameters $\boldsymbol{\theta}^{(1:K)}$. Although parameters differ from system to system, they are not entirely independent -- they reflect variations within a common population, for example due to manufacturing or environmental conditions. To represent this structured variability, 
each parameter $\boldsymbol{\theta}^{(k)}$ is generated from a common population,
% \begin{equation}
% p(\boldsymbol{\theta}^{(k)}|\boldsymbol{\phi}) \sim \mathcal{N}(\boldsymbol{\theta}^{(k)};\boldsymbol{\mu}_{\phi},\boldsymbol{\tau}_{\phi}\mathbf{I}), \label{hierarchical}
% \end{equation}
\begin{equation}
    \label{hierarchical}
    \boldsymbol{\theta}^{(k)} | \boldsymbol{\phi} \sim p(\boldsymbol{\theta}^{(k)}|\boldsymbol{\phi}),
\end{equation}
with hyperparameters $\boldsymbol{\phi}$ governing the population distribution, known as the hyperprior,
% \begin{equation}
%     \label{eq:hyperprior_mean}
%     p(\boldsymbol{\phi}) \sim \prod_{i=1}^{N}\mathcal{N}( \mu_{\phi_i};\mu_i,\tau_i) \times \prod_{i=1}^{N}\mathcal{N}( \log(\tau_{\phi_i});a_i,b_i),
% \end{equation}
\begin{equation}
    \label{hyperprior}
    \boldsymbol{\phi} \sim p(\boldsymbol{\phi}).
\end{equation}
% where $i$ indexes unknown parameters of interest. Parameters $\mu_i$, $\tau_i$, $a_i$, and $b_i$ are known values, encoding prior beliefs about the hyperprior distribution. Logarithm of $\boldsymbol{\tau}_{\phi}$ is used to ensure that the variance in the hyperprior is always positive.
Observations $\boldsymbol{y}^{(k)}$ are then obtained from each $\boldsymbol{\theta}^{(k)}$ through
\begin{equation}
    \label{hierarhical_obser}
    \boldsymbol{y}^{(k)} | \boldsymbol{\theta}^{(k)}, \boldsymbol{\phi} \sim p(\boldsymbol{y}^{(k)} | \boldsymbol{\theta}^{(k)}, \boldsymbol{\phi}).
\end{equation}
The structure of the hierarchical Bayesian model is illustrated in Figure \ref{fig:hier_bayes}.

We now detail the factorization of some key distributions using the hierarchical Bayesian framework. The joint prior distribution factorizes into the product of the hyperprior and individual conditional priors,
\begin{equation}
    \label{eq:joint_prior}
    p(\boldsymbol{\theta}^{(1:K)},\boldsymbol{\phi}) = p(\boldsymbol{\phi})\prod_{k=1}^{K}p(\boldsymbol{\theta}^{(k)}|\boldsymbol{\phi}).
\end{equation}
Consequently, the joint posterior distribution of the parameters and hyperparameters, conditioned on the observations $\mathbf{y}^{(1:K)}$, is expressed as
\begin{equation}
    p(\boldsymbol{\theta}^{(1:K)},\boldsymbol{\phi}|\boldsymbol{y}^{(1:K)})=\frac{p(\boldsymbol{\phi})\prod_{k=1}^{K}p(\mathbf{y}^{(k)}|\boldsymbol{\theta}^{(k)})p(\boldsymbol{\theta}^{(k)}|\boldsymbol{\phi}) }{p(\boldsymbol{y}^{(1:K)})},
\end{equation}
which is the distribution that we want to sample from. Next, we will discuss the sampling scheme used in our method.

\subsection{Ensemble MALA}\label{subsec2}

Metropolis-Adjusted Langevin Algorithm (MALA) augments standard Metropolis-Hastings sampling with gradient-informed proposals based on discretizations of Langevin diffusion. Given a target density $p$, the associated Langevin SDE is
\begin{align}
    \mathrm{d}X_t = \nabla \log p(X_t) \mathrm{d}t + \sqrt{2}\mathrm{d}W_t,
\end{align}
where $W_t$ is a standard $d$-dimensional Brownian motion, then the stochastic process $X_t$ has an invariant distribution with density $p$. After an Euler-Maruyama discretization, we obtain
\begin{equation}
    \label{eq:langevin_disc}
    X_{\mathsf{t}+1} = X_\mathsf{t} + \gamma \nabla _{X_\mathsf{t}} \log p (X_\mathsf{t}) + \sqrt{2\gamma}  \epsilon_\mathsf{t}, \ \ \epsilon_\mathsf{t} \sim \mathcal{N}(0,I),
\end{equation}
where $\gamma$ is the step size and MALA has a further Metropolis accept-reject step. By evolving the state $X$ following \eqref{eq:langevin_disc}, one obtains samples asymptotically distributed according to the target stationary distribution. However, a single Langevin chain may converge slowly in high-dimensional or multimodal posteriors, ensemble MCMC strategies accelerate convergence by evolving multiple interacting chains and integrating ensemble-based covariance information.

The use of ensembles was originally developed in ensemble Kalman filter (EnKF) literature \citep{evensen2009data,stuart2014data}, where an ensemble of states is used to approximate the covariance of the underlying distribution of system states. These ideas were later extended to inverse problems through the ensemble Kalman inversion (EKI) framework \citep{iglesias2013ensemble}. The development of ensemble Kalman sampler (EKS) \citep{garbuno2020interacting} is a noisy extension of EKI that incorporates ideas from Langevin diffusions while avoiding the need to evaluate gradients of the forward model. Its core mechanism relies on an interacting system of multiple Langevin chains, where each chain evolves according to a preconditioned Langevin dynamics and the empirical ensemble covariance serves as the preconditioner. A similar interacting particle framework was recently adopted in the Interacting Particle Langevin Algorithm (IPLA) \citep{akyildiz2025interacting} to address maximum marginal likelihood estimation, which has recently been applied to the PDE inversion setting \citep{glyn2025statistical}. 

Adopting this perspective of interacting chains, our ensemble MALA employs an ensemble of $M$ parallel preconditioned Langevin chains. Each chain evolves the state vector $X_\mathsf{t}^m$, for chain index $m = 1, \ldots, M$. The preconditioning matrix $\mathbf{C}_\mathsf{t}$ is adaptively computed as the empirical covariance across chains at iteration $\mathsf{t}$, thereby aligning the sampling dynamics with the local geometry of the posterior distribution. Such pre-conditioning can drastically improve sampling stability and efficiency. Each preconditioned Langevin chain is
\begin{equation}
    \label{eq:ensembleMALA}
    X_{\mathsf{t}+1}^m = X_\mathsf{t}^m + \gamma \mathbf{C}_\mathsf{t} \nabla_{X_\mathsf{t}^m} \log p(X_\mathsf{t}^m) + \sqrt{2\gamma \mathbf{C}_\mathsf{t}}\epsilon_{\mathsf{t}}^m,
\end{equation}
and the preconditioning matrix $\mathbf{C}_\mathsf{t}$ is given by
\begin{equation}
    \label{eq:preconditioner}
    \mathbf{C}_\mathsf{t}=\frac{1}{M}\sum_{m=1}^{M}(X_\mathsf{t}^m - \overline{X_\mathsf{t}} )\otimes (X_\mathsf{t}^m - \overline{X_\mathsf{t}} ),
\end{equation}
where $\overline{X_\mathsf{t}}$ is the empirical mean of states at iteration $\mathsf{t}$. To correct for the bias introduced by the time-discretization of the Langevin SDE, a Metropolis accept-reject step is applied. The proposal $X_{\mathsf{t}+1}^m$ is accepted with probability
\begin{equation}
    \label{eq:accept_probability}
    a = \min \left \{ \frac{ p (X_{\mathsf{t}+1}^m)q (X_\mathsf{t}^m|X_{\mathsf{t}+1}^m)}{p (X_{\mathsf{t}}^m)q (X_{\mathsf{t}+1}^m|X_{\mathsf{t}}^m)},1\right \},
\end{equation}
and the proposal distribution $q(\cdot|\cdot)$ could be derived from \eqref{eq:ensembleMALA} and is
\begin{equation}
    \label{eq:accept_probability}
    q(X_{\mathsf{t+1}}|X_{\mathsf{t}}) \propto \exp \left (-\frac{1}{4\gamma} \left \| X_{\mathsf{t+1}}-X_\mathsf{t}-\gamma \mathbf{C_\mathsf{t}} \nabla_{X_\mathsf{t}} \log p (X_{\mathsf{t}})  \right \| ^2 _{\mathbf{C_\mathsf{t}}} \right ).
\end{equation}
In the next session, we will discuss our proposed methodology built on the mentioned backgrounds.

\section{Methodology}\label{sec4}

This section details how hierarchical parameter inference and maximum likelihood closure are jointly optimized. Additionally, we present the bilevel optimization framework to jointly train surrogate models, thereby accelerating the inverse problem. Finally, the specific surrogate model structures and their respective training methodologies are discussed in detail.

% \subsection{Inverse Problems using Numerical Solvers}\label{subsec3}
\subsection{Hierarchical Inverse Problem and Maximum Likelihood Closure}\label{subsec3}

We begin by describing how the inverse problem can be solved using a traditional numerical forward solver, without introducing surrogate models. This section corresponds to our contributions \ref{con:c1} and \ref{con:c2}. The forward model is given in \eqref{eq:general_system}, and our objective is to infer both physical parameters $\boldsymbol{\theta}^{(1:K)}$ and $\boldsymbol{\phi}$, and nonlinear closure $f$ from sparse and noisy observations $\mathbf{y}^{(1:K)}$. As discussed in Section \ref{subsec1}, parameters from different physical systems are modeled using a hierarchical Bayesian framework, and we assume each $\boldsymbol{\theta}^{(k)}$ is drawn from a common Gaussian distribution,
\begin{equation}
p(\boldsymbol{\theta}^{(k)}|\boldsymbol{\phi}) = \mathcal{N}(\boldsymbol{\theta}^{(k)};\boldsymbol{\mu}_{\phi},\boldsymbol{\tau}_{\phi}\mathbf{I}), \label{hierarchical}
\end{equation}
with $\boldsymbol{\phi} = \{\boldsymbol{\mu}_\phi,\boldsymbol{\tau}_\phi \}$ and the hyperprior
\begin{equation}
    \label{eq:hyperprior_mean}
    p(\boldsymbol{\phi}) = p(\boldsymbol{\mu}_\phi)p(\boldsymbol{\tau}_\phi) = \prod_{i=1}^{P}\mathcal{N}( \mu_{\phi_i};m_\mu,s_\mu)\mathcal{N}( \log(\tau_{\phi_i});m_\tau,s_\tau),
\end{equation}
where $i$ indexes unknown parameters of interest and there are $P$ unknown parameters in each system. Parameters $m_\mu$, $s_\mu$, $m_\tau$, and $s_\tau$ are known values, encoding prior beliefs about the hyperprior distribution. Working with the logarithm of $\boldsymbol{\tau}_{\phi}$ corresponds to a log-normal distribution and ensures that $\tau_{\phi_i
}$ in the hyperprior is always positive. 
% In this setting, $\boldsymbol{\theta}^{(1:K)}$ and $\boldsymbol{\phi}$ are both unknown parameters to be inferred and the joint prior distribution is
% \begin{equation}
%     p(\boldsymbol{\theta}^{(1:K)},\boldsymbol{\phi}) = p(\boldsymbol{\phi})\prod_{k=1}^{K}p(\boldsymbol{\theta}^{(k)}|\boldsymbol{\phi}).
% \end{equation}

To represent the unknown nonlinear closure, we employ a neural network $f^{\alpha}$ with trainable parameters $\alpha$. Observations of each system are obtained via \eqref{eq:observation_operation}, so the likelihood of observations from system $k$ takes the form
\begin{equation}
    \label{eq:likelihood}
p(\mathbf{y}^{(k)}|\boldsymbol{\theta}^{(k)},\alpha) = \mathcal{N}(\mathbf{y}^{(k)};G^{(k)}(\boldsymbol{\theta}^{(k)},\alpha),\sigma^2\mathbf{I}),
\end{equation}
where $G^{(k)} = g^{(k)} \circ F^\dagger$, denoting the operator that maps $(\boldsymbol{\theta}^{(k)},\alpha)$ to corresponding observations. We simplify the input as $\alpha$ here to represent $f^{\alpha}$. Depending on the structure of governing equations, the parameter-to-solution operator $F^\dagger$ may correspond to a leapfrog integrator or Euler time‐stepping scheme for dynamical systems, or a Gauss–Newton solver for nonlinear stationary problems.

To learn the closure model $f^\alpha$, we maximize the marginal likelihood of the observational data $\mathbf{y}^{(1:K)}$ with respect to $\alpha$. 
However, due to the presence of unknown latent variables $\boldsymbol{\theta}^{(1:K)}$ and $\boldsymbol{\phi}$, the log marginal is
\begin{align}
    \label{eq:log_marginal}
    \log p(\mathbf{y}^{(1:K)} | \alpha) 
    % &= \sum_{k=1}^{K}  \log p(\mathbf{y}^{(k)}|\alpha), \\
    % &= \sum_{k=1}^{K} \log \left ( \int p(\mathbf{y}^{(k)}|\boldsymbol{\theta}^{(k)},\alpha)p(\boldsymbol{\theta}^{(k)},\boldsymbol{\phi}|\alpha) d(\boldsymbol{\theta}^{(k)},\boldsymbol{\phi})\right ) .
    &= \log \left ( \int p(\mathbf{y}^{(1:K)}|\boldsymbol{\phi},\alpha) p(\boldsymbol{\phi}|\alpha) d\boldsymbol{\phi} \right ), \nonumber \\
    &= \log \left ( \int \left[ \prod_{k=1}^K \int p(\mathbf{y}^{(k)}|\boldsymbol{\theta}^{(k)}, \alpha) p(\boldsymbol{\theta}^{(k)}|\boldsymbol{\phi}) d\boldsymbol{\theta}^{(k)} \right] p(\boldsymbol{\phi}|\alpha) d\boldsymbol{\phi} \right ),
\end{align}
which is both analytically and computationally intractable.
Therefore, we consider another useful quantity arising in parameter estimation problems, which is the gradient of the log marginal likelihood with respect to $\alpha$. Following Fisher's identity \citep{douc2014nonlinear},
\begin{equation}
    \label{eq:fishers}
    \nabla_\alpha \log p(\mathbf{y}^{(1:K)} | \alpha) = \mathbb{E}_{p(\boldsymbol{\theta}^{(1:K)},\boldsymbol{\phi}|\mathbf{y}^{(1:K)},\alpha)}\left [ \nabla_\alpha \log p(\mathbf{y}^{(1:K)},\boldsymbol{\theta}^{(1:K)} ,\boldsymbol{\phi}| \alpha)\right ].
\end{equation}
The expectation is taken with respect to the joint posterior distribution $p(\boldsymbol{\theta}^{(1:K)},\boldsymbol{\phi}|\mathbf{y}^{(1:K)},\alpha)$, which can be factorized as 
\begin{align}
    \label{eq:joint_posterior}
    p(\boldsymbol{\theta}^{(1:K)},\boldsymbol{\phi}|\boldsymbol{y}^{(1:K)},\alpha) 
    &= \frac{p(\mathbf{y}^{(1:K)},\boldsymbol{\theta}^{(1:K)} ,\boldsymbol{\phi}| \alpha)}{p(\boldsymbol{y}^{(1:K)}|\alpha)}, \nonumber \\
    &=\frac{p(\boldsymbol{\phi})\prod_{k=1}^{K}p(\mathbf{y}^{(k)}|\boldsymbol{\theta}^{(k)},\alpha)p(\boldsymbol{\theta}^{(k)}|\boldsymbol{\phi}) }{p(\boldsymbol{y}^{(1:K)}|\alpha)}.
\end{align}
The term inside the expectation brackets, $p(\mathbf{y}^{(1:K)},\boldsymbol{\theta}^{(1:K)} ,\boldsymbol{\phi}| \alpha)$, represents the joint probability density and is the numerator in \eqref{eq:joint_posterior}.
% \begin{equation}
%     \label{eq:jointdistribution}
%     p(\boldsymbol{y}^{(1:K)},\boldsymbol{\theta}^{(1:K)}, \boldsymbol{\phi}|\alpha) = p(\boldsymbol{\phi})\prod_{k=1}^{K}p(\mathbf{y}^{(k)}| \boldsymbol{\theta}^{(k)},\alpha)p(\boldsymbol{\theta}^{(k)}|\boldsymbol{\phi}).
% \end{equation}
% Since the posterior distribution in \eqref{eq:joint_posterior} is intractable, the expectation required for gradient computation cannot be derived analytically, so
We use samples from~\eqref{eq:joint_posterior} to approximate the expectation in \eqref{eq:fishers}. 

% We adopt an iterative scheme on posterior sampling and parameter update, where one Langevin sampling step and one gradient descent step on $\alpha$ are iteratively conducted.
To learn $\alpha$ through~\eqref{eq:fishers}, we interleave the sampling of~\eqref{eq:joint_posterior} using ensemble MALA, and gradient descent on $\alpha$.
Parameters are initialized from $\alpha_0$ and an ensemble of $M$ MALA chains is maintained, with state vector starting from $X_\mathsf{0}^{1:M} = (\boldsymbol{\theta}^{(1:K)}, \boldsymbol{\phi})_\mathsf{0}^{1:M}$. 
At iteration $\mathsf{t}$, given the current parameter estimate $\alpha_\mathsf{t}$, we perform a single step of Langevin sampling across all chains. Utilizing the ensemble MALA procedure detailed in Section \ref{subsec2}, samples $X_{\mathsf{t}+1}^{1:M}$ are drawn from the current posterior distribution $p(\boldsymbol{\theta}^{(1:K)},\boldsymbol{\phi}|\boldsymbol{y}^{(1:K)},\alpha_\mathsf{t})$.
% At iteration $\mathsf{t}$, where the current parameter estimate is $\alpha_\mathsf{t}$, Langevin sampling proceeds one step for each chain, obtaining samples $X_{\mathsf{t}+1}^{1:M}$ from the current posterior distribution $p(\boldsymbol{\theta}^{(1:K)},\boldsymbol{\phi}|\boldsymbol{y}^{(1:K)},\alpha_\mathsf{t})$ using ensemble MALA as in Section \ref{subsec2}.
% , evolving the chains for one further step to
% \begin{equation}
%     X_\mathsf{t+1}^{1:M} = (\boldsymbol{\theta}^{(1:K)}, \boldsymbol{\phi})_\mathsf{t+1}^{1:M} .
% \end{equation}
These samples from $M$ chains are then used to approximate the gradient in \eqref{eq:fishers} to obtain $\alpha_{\mathsf{t+1}}$, where the gradient for update is 
\begin{equation}
    \label{eq:alpha_gradient}
    \nabla_\alpha \log p(\mathbf{y}^{(1:K)} | \alpha_\mathsf{t}) \approx  \sum_{m=1}^{M} \nabla_\alpha \log p(\boldsymbol{y}^{(1:K)},(\boldsymbol{\theta}^{(1:K)}, \boldsymbol{\phi})^m_\mathsf{t+1}|\alpha_\mathsf{t}).
\end{equation}
To be compatible with standard deep learning frameworks (where one defines a function to be differentiated using automatic differentiation), we define the objective function for learning $\alpha$ as
\begin{equation}
    \label{eq:alpha_loss}
    \mathcal{L}_\mathsf{LML}(\alpha) = - \sum_{m=1}^{M} \log p(\boldsymbol{y}^{(1:K)},(\boldsymbol{\theta}^{(1:K)}, \boldsymbol{\phi})^m_\mathsf{t+1}|\alpha),
\end{equation}
where unnecessary constants are dropped.

Upon convergence of the training loss $\mathcal{L}_{\mathsf{LML}}$ after $\mathsf{T}$ iterations, we initiate a post-training sampling phase. We fix the neural network parameters $\alpha_{\mathsf{T}}$ and the preconditioner $\mathbf{C_{\mathsf{T}}}$, and then advance the MALA chains to sample from the resulting stationary posterior distribution $p(\boldsymbol{\theta}^{(1:K)},\boldsymbol{\phi}|\boldsymbol{y}^{(1:K)},\alpha_{\mathsf{T}})$. This yields the final posterior samples for the system parameters $\boldsymbol{\theta}^{(1:K)}$ and hyperparameters $\boldsymbol{\phi}$, alongside the learned closure model defined by $\alpha_{\mathsf{T}}$.

\subsection{Joint Training of Forward Surrogate}\label{subsec4}

Because ensemble MALA requires a large number of forward simulations, and each proposal step involves evaluating gradients of the forward solve, using traditional numerical solvers can become computationally prohibitive. To address this limitation, we replace the numerical forward solver with a surrogate model to be concurrently learned in the inference scheme. This answers contribution \ref{con:c3}. In this work, we consider both FNOs and PINNs as surrogate architectures, although other differentiable models could be used as well. 

We denote the surrogate forward model by $F^{\beta}$. Different from the numerical forward solver in \eqref{eq:general_system}, surrogate model cannot take high-dimensional neural network parameters $\alpha$ directly as input. Therefore, $F^\beta$ approximates the numerical solver evaluated at $f^\alpha \approx f$, denoted by
\begin{equation}
    \label{eq:surrogate_approximate}
    F^\dagger (\boldsymbol{\theta}^{(k)},\alpha) \approx F^{\beta}(\boldsymbol{\theta}^{(k)}) = \hat{\mathbf{u}}^{(k)}.
\end{equation}
We expect this approximation to be accurate in regions of high probability under the posterior distribution of $\boldsymbol{\theta}^{(k)}$ conditioned on observational data, with $k \in \{ 1,...,K\}$, and  $(\boldsymbol{\theta}^{(1:K)},\boldsymbol{\phi} )\sim p(\boldsymbol{\theta}^{(1:K)},\boldsymbol{\phi}|\bfy^{(1:K)}, \alpha)$.

Although $F^{\beta}$ does not take $\alpha$ directly as input, the influence of $\alpha$ enters indirectly through the surrogate training loss $\mathcal{L}_{\text{Surrogate}}(\alpha,\beta)$, with its form discussed later in Section \ref{subsec5}. In the following sections, we first introduce how surrogate training is integrated with the inverse problem through a bilevel optimization framework, enabling joint learning of the surrogate and unknown physics. We then describe the forms of different loss functions used to train these surrogate models. 

\begin{figure}[t!]
    \centering
    \resizebox{\linewidth}{!}{%
    \begin{tikzpicture}[
        >=Stealth,
        font=\sffamily\small,
        data_box/.style={rectangle, rounded corners=12pt, fill=gray!12, draw=gray!30, thick, minimum width=15cm, minimum height=3cm},
        mala_box/.style={rectangle, rounded corners=12pt, fill=blue!8, draw=blue!20, thick, minimum width=11cm, minimum height=3.8cm},
        surro_box/.style={rectangle, rounded corners=12pt, fill=green!8, draw=green!20, thick, minimum width=8.4cm, minimum height=6.2cm},
        clos_box/.style={rectangle, rounded corners=12pt, fill=orange!8, draw=orange!20, thick, minimum width=8.4cm, minimum height=6.2cm},
        arrow/.style={->, thick, draw=black!70},
        dashed_arrow/.style={->, thick, dashed, draw=black!70},
        label/.style={fill=white, font=\footnotesize, inner sep=2pt, rounded corners=2pt},
        nn_node/.style={circle, draw=orange!80!black, fill=white, inner sep=0pt, minimum size=0.4cm, thick}
    ]

    % ==========================================
    % 1. Hierarchical Setup & Observation Data
    % ==========================================
    \node[data_box] (data) at (0, 7.5) {};
    
    \draw[gray!30, thick] (-2.5, 6.1) -- (-2.5, 8.9);
    
    % Hierarchical Setup
    \begin{scope}[shift={(-5.0, 7.5)}]
        \node[font=\bfseries, text=gray!80!black] at (0, 1.0) {Hierarchical Setup};
        \node[font=\large] (phi) at (-1.5, -0.2) {$\boldsymbol{\phi}$};
        \node[font=\large] (theta) at (-0.1, -0.2) {$\boldsymbol{\theta}^{(1:K)}$};
        \node[font=\large] (y) at (1.5, -0.2) {$\mathbf{y}^{(1:K)}$};
        
        \draw[->, very thick, gray!80!black] (phi) -- (theta);
        \draw[->, very thick, gray!80!black] (theta) -- (y);
    \end{scope}

    % Observation Data
    \begin{scope}[shift={(2.8, 7.5)}]
        \node[font=\bfseries, text=gray!80!black] at (0, 1.0) {Observation Data};
        
        \node[inner sep=0pt] (sys1) at (-2.5, -0.3) {
            \includegraphics[width=1.7cm, height=1.7cm]{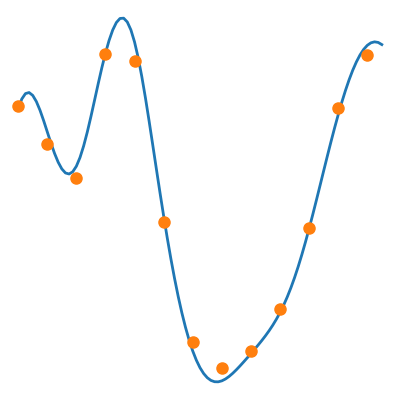}
        };
        
        \node[inner sep=0pt] (sys2) at (0, -0.3) {
            \includegraphics[width=1.7cm, height=1.7cm]{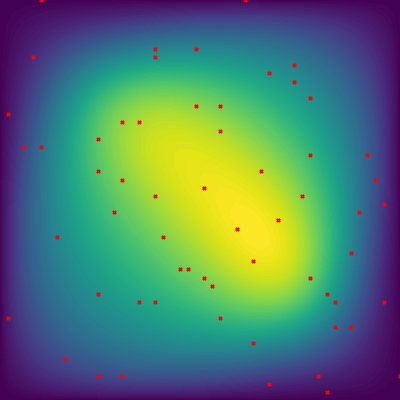}
        };
        
        \node[inner sep=0pt] (sys3) at (2.5, -0.3) {
            \includegraphics[width=1.7cm, height=1.7cm]{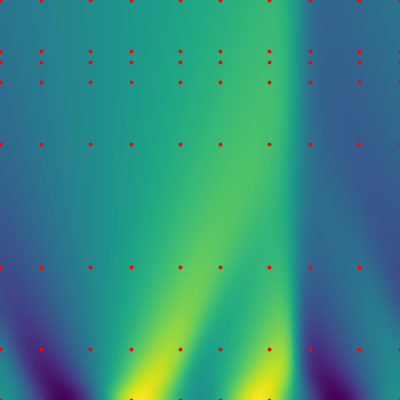}
        };
    \end{scope}

    % ==========================================
    % 2. Posterior Sampling (MALA)
    % ==========================================
    \node[mala_box] (mala) at (0, 2.5) {};
    \node[anchor=north, text=blue!80!black, font=\bfseries] at (mala.north) {\vspace{0.15cm}1. Hierarchical Posterior Sampling};
    
    \begin{scope}[shift={(mala.center)}, yshift=-0.2cm]
        \node[font=\normalsize, text=blue!90!black] at (0, 1.0) {$X_{\mathsf{t}}^m \xrightarrow{\text{ Langevin }} X_{\mathsf{t}+1}^m$};
        
        \draw[blue!20, thick, fill=blue!4] (0,-0.1) ellipse (1.6cm and 0.7cm);
        \draw[blue!40, thick, fill=blue!12] (0.2,-0.1) ellipse (0.8cm and 0.35cm);
        
        \fill[red] (-1.0, -0.4) circle (2pt); \draw[->, red, thick] (-1.0, -0.4) -- (-0.6, -0.2);
        \fill[red] (1.1, 0.2) circle (2pt);   \draw[->, red, thick] (1.1, 0.2) -- (0.7, 0.0);
        \fill[red] (-0.5, 0.4) circle (2pt);  \draw[->, red, thick] (-0.5, 0.4) -- (-0.2, 0.1);
        \fill[red] (0.8, -0.5) circle (2pt);  \draw[->, red, thick] (0.8, -0.5) -- (0.5, -0.3);
        
        \node[font=\normalsize, text=blue!80!black] at (0, -1.15) {$X_{t+1}^m = (\boldsymbol{\theta}^{(1:K)}, \boldsymbol{\phi})_{t+1}^m \sim p(\cdot \mid \mathbf{y}^{(1:K)}, \alpha_\mathsf{t}, \beta_{\mathsf{t} N})$};
    \end{scope}

    % ==========================================
    % 3. Surrogate Training (Lower Level)
    % ==========================================
    \node[surro_box] (surro) at (-5.0, -4.5) {};
    \node[anchor=north, text=green!50!black, font=\bfseries] at (surro.north) {\vspace{0.15cm}2. Lower Level (Surrogate Training)};
    
    \begin{scope}[shift={(surro.center)}, yshift=0.3cm]
        \node[font=\Large] (in_theta) at (-2.5, -0.05) {$\boldsymbol{\theta}^{(1:K)}$};
        
        \node[rectangle, draw=green!60!black, fill=white, thick, rounded corners=4pt, minimum width=2.4cm, minimum height=2.5cm, align=center] (fno_box) at (0, -0.05) {\normalsize \textbf{PINN / FNO} \\[0.15cm] \normalsize ($F^\beta$)};
        
        \draw[->, thick, green!60!black] (in_theta) -- (in_theta -| fno_box.west);
        
        \node[font=\Large] (out_u) at (2.5, -0.05) {$\hat{\mathbf{u}}^{(1:K)}$};
        \draw[->, thick, green!60!black] (fno_box.east) -- (out_u);
        
        % Loss 
        \node[rectangle, draw=green!60!black, fill=white, thick, rounded corners=3pt, font=\small, inner sep=4pt] (loss_surro) at (2.5, -2.4) {$\mathcal{L}_{\mathsf{Surrogate}}(\alpha, \beta)$};
        \draw[->, thick, gray!80] (out_u.south) -- (loss_surro.north);
        
        \draw[->, dashed, thick, green!60!black] (loss_surro.south) -- ++(0, -0.3) coordinate (turn_surro) -- node[below, font=\scriptsize, text=green!60!black] {Update $\beta$ ($N$ iter)} (fno_box.south |- turn_surro) -- (fno_box.south);
    \end{scope}

    % ==========================================
    % 4. Closure Learning (Upper Level)
    % ==========================================
    \node[clos_box] (clos) at (5.0, -4.5) {};
    \node[anchor=north, text=orange!80!black, font=\bfseries] at (clos.north) {\vspace{0.15cm}3. Upper Level (Closure Learning)};
    
    \begin{scope}[shift={(clos.center)}, yshift=0.3cm]
        \node[font=\Large] (in_u) at (-2.5, 0) {$u$};
        \node[font=\Large] (out_f) at (2.5, 0) {$f^\alpha(u)$};

        \foreach \i/\y in {1/0.8, 2/0, 3/-0.8} {
            \node[nn_node] (h1_\i) at (-1.2, \y) {};
            \node[nn_node] (h2_\i) at (-0.4, \y) {};
            \node[font=\normalsize, text=orange!80!black] (dots_\i) at (0.4, \y) {$\dots$};
            \node[nn_node] (h3_\i) at (1.2, \y) {};
        }

        \node[rectangle, draw=orange!80!black, dashed, thick, rounded corners=6pt, inner sep=6pt, fit=(h1_1) (h3_1) (h1_3) (h3_3)] (mlp_box) {};
        
        \node[font=\normalsize, text=orange!80!black, fill=orange!8, inner sep=2pt] at ([yshift=0.05cm]mlp_box.north) {\textbf{MLP} ($f^\alpha$)};
        
        \foreach \i in {1,2,3} \draw[->, thick, orange!80!black] (in_u) -- (h1_\i);
        
        \foreach \ia in {1,2,3} {
            \foreach \ib in {1,2,3} {
                \draw[thick, orange!80!black!70] (h1_\ia) -- (h2_\ib); 
            }
        }
        
        \foreach \i in {1,2,3} \draw[thick, orange!80!black] (h2_\i) -- (dots_\i);

        \foreach \i in {1,2,3} \draw[thick, orange!80!black] (dots_\i) -- (h3_\i);
        
        \foreach \i in {1,2,3} \draw[->, thick, orange!80!black] (h3_\i) -- (out_f);

        \node[rectangle, draw=orange!80!black, fill=white, thick, rounded corners=3pt, font=\small, inner sep=4pt] (loss_clos) at (2.5, -2.4) {$\mathcal{L}_{\mathsf{LML}}(\alpha, \beta^*(\alpha))$};
        \draw[->, thick, gray!80] (out_f.south) -- (loss_clos.north);
        
        \draw[->, dashed, thick, orange!80!black] (loss_clos.south) -- ++(0, -0.3) coordinate (turn_clos) -- node[below, font=\scriptsize, text=orange!80!black] {Update $\alpha$ (1 iter)} (mlp_box.south |- turn_clos) -- (mlp_box.south);
    \end{scope}
    
    % Data -> MALA 
    \draw[arrow] (data.south) -- node[right, pos=0.3, font=\normalsize, text=gray!80!black] {$\mathbf{y}^{(1:K)}$} (mala.north);
    
    % MALA -> Surrogate 
    \draw[arrow] ([xshift=-5.0cm]mala.south) -- node[left, align=center, font=\normalsize, text=black!80] {One random chain \\ $X_{\mathsf{t}+1}^{m}$} (surro.north);

    % MALA -> Closure 
    \draw[arrow] ([xshift=5.0cm]mala.south) -- node[right, align=center, font=\normalsize, text=black!80] {All chains \\ $X_{\mathsf{t}+1}^{1:M}$} (clos.north);

    % Surrogate -> Closure 
    \draw[arrow] (surro.east) -- node[above, font=\normalsize, text=black!90] {$\beta^*(\alpha)$} (clos.west);

    \draw[dashed_arrow, rounded corners=8pt] (clos.east) -- ++(1.0,0) |- node[right, pos=0.25] {Updated $\alpha_{\mathsf{t}+1}$} (mala.east);
    \draw[dashed_arrow, rounded corners=8pt] (surro.west) -- ++(-1.0,0) |- node[left, pos=0.25] {Updated $\beta_{(\mathsf{t}+1) N}$} (mala.west);

    \begin{scope}[on background layer]
        \node[fill=gray!3, rounded corners=12pt, draw=gray!30, dashed, thick, inner sep=2.5em, fit=(mala) (surro) (clos)] (bg) {};
        \node[below right, text=gray!60!black, font=\bfseries] at (bg.north west) {Iteration $\mathsf{t} \rightarrow \mathsf{t}+1$};
    \end{scope}

    \end{tikzpicture}%
    }
    \caption{\textbf{Overview of the proposed framework.} The algorithm alternates between three main stages: (1) \textbf{Hierarchical posterior sampling} using ensemble MALA to update parameter chains, (2) \textbf{Lower-level surrogate training} optimizing the surrogate model $F^\beta$ parameterized by $\beta$ over $N$ iterations, where the loss function $\mathcal{L}_{\mathsf{Surrogate}}$ could take the forms in \eqref{eq:supervised_loss}, \eqref{eq:fno_physics_loss} and \eqref{eq:PINNs_loss}, and (3) \textbf{Upper-level closure learning} optimizing the closure model $f^\alpha$ parameterized by $\alpha$ to discover the unknown nonlinear dynamics, where the loss function $\mathcal{L}_{\mathsf{LML}}$ follows \eqref{eq:alpha_loss_2}.}
    \label{fig:framework}
\end{figure}
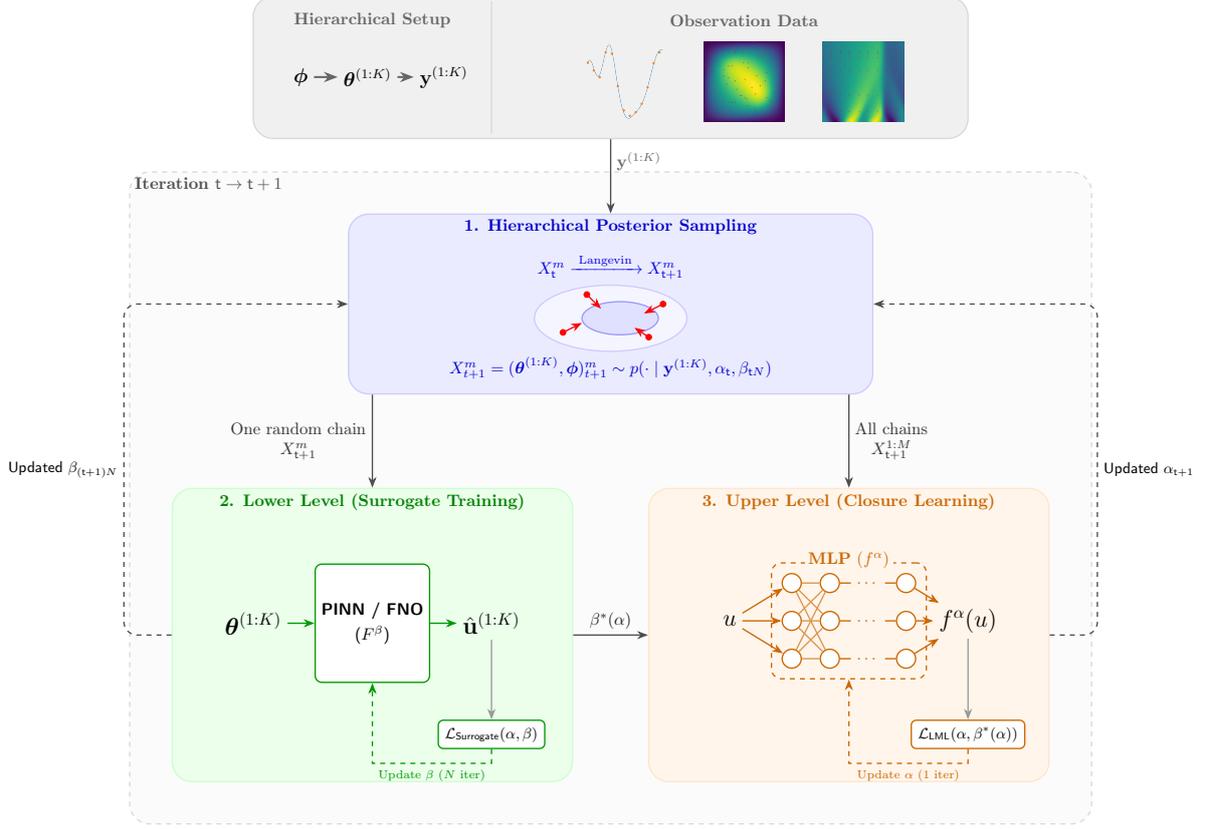

\subsubsection{Bilevel Update Scheme}\label{subsec6}

The joint learning of forward and inverse problems naturally leads to a bilevel optimization structure, in which two coupled subproblems need to be solved simultaneously. Bilevel optimization \citep{sinha2017review} refers to problems where the solution of an upper-level objective depends on the optimal solution of a lower-level problem. In this work, the bilevel problem takes the form
\begin{equation}
    \label{eq:bilevel}
    \min_{\alpha}\; \mathcal{L}_\mathsf{LML}(\alpha,\beta^\ast(\alpha))
    \qquad\text{s.t.}\qquad
    \beta^*(\alpha) = \underset{\beta}{\mathrm{arg\,min}} \, \mathcal{L}_\mathsf{Surrogate}(\alpha, \beta),
\end{equation}
where $\mathcal{L}_\mathsf{LML}$ represents the upper-level objective (maximizing marginal likelihood) and $\mathcal{L}_\mathsf{Surrogate}$ represents the lower-level objective (training the surrogate). This formulation entails an iterative interplay between the two levels. At each step, the lower-level variables $\beta$ are first optimized conditioned on the current $\alpha$. The closure parameters $\alpha$ are then updated by differentiating through the upper-level objective, while accounting for the implicit dependence of the optimal surrogate $\beta^{\ast}$ on $\alpha$. This mechanism effectively propagates gradient information from the inner surrogate training loop to the outer closure learning loop. The bilevel structure arises from the mutual dependency between the two components:

\begin{itemize}
\item \textbf{Upper-Level (Closure Learning):} Optimizing closure parameters $\alpha$ requires evaluating the marginal likelihood of the data. This evaluation relies on the surrogate forward model $F^{\beta^\ast(\alpha)}$ to map parameters $\boldsymbol{\theta}$ to observations $\mathbf{y}$.
\item \textbf{Lower-Level (Surrogate Training):} Training the surrogate model parameterized by $\beta$ requires knowledge of the governing equations, which includes the form of current closure model $f^\alpha$. Thus, the optimal surrogate parameters $\beta^*$ are implicitly dependent on $\alpha$.
\end{itemize}

Next, we detail the complete training scheme of our methodology. Each iteration consists of three phases: one sampling step, $N$ optimization steps on $\beta$, and one update step on $\alpha$. At iteration $\mathsf{t}$, the sampling step is first conducted using the ensemble MALA sampler, obtaining
\begin{equation}
\label{eq:bilevel_sampling}
(\boldsymbol{\theta}^{(1:K)}, \boldsymbol{\phi})^{1:M}_{\mathsf{t+1}} \sim p(\boldsymbol{\theta}^{(1:K)}, \boldsymbol{\phi}|\boldsymbol{y}^{(1:K)},\alpha_{\mathsf{t}},\beta_{\mathsf{t} N}).
\end{equation}
Note that the efficient surrogate is used here to compute the gradient in Langevin sampling and acceptance probability, avoiding expensive calls of the numerical solver. Additionally, the gradient propagation is blocked before the sampling stage, and the optimization of $\alpha$ and $\beta$ does not involve differentiation through the MALA sampling trajectory.

Then, $N$ lower-level optimization steps are performed to update $\beta$. To reduce computational cost, the loss is evaluated using samples from only one random MALA chain $(\boldsymbol{\theta}^{(1:K)}, \boldsymbol{\phi})^p_\mathsf{t+1}$, where the chain index $p$ is drawn from $\mathrm{Uniform}\{1,\dots,M\}$. $N$ gradient descent steps on $\beta$ are performed to minimize $\mathcal{L}_\mathsf{Surrogate}(\alpha,\beta)$, with its specific form discussed later in Section \ref{subsec5}. 

Finally, at the upper level, closure parameters $\alpha$ are updated with surrogate parameters fixed at their optimized state $\beta^\ast(\alpha) = \beta_{\mathsf{(t+1)}N}$. The objective function is similar to what is mentioned in Section \ref{subsec3}. However, as previously noted, the surrogate model does not explicitly take $\alpha$ as input. Therefore, some representations need to be changed. The original likelihood distribution in \eqref{eq:likelihood} is reformulated to
\begin{equation}
    \label{eq:likelihood_new}
    p(\mathbf{y}^{(k)}|\boldsymbol{\theta}^{(k)},\beta^*(\alpha)) = \mathcal{N}(\mathbf{y}^{(k)};g^{(k)} \circ F^{\beta^*(\alpha)}(\boldsymbol{\theta}^{(k)}),\sigma^2\mathbf{I}).
    %\tag{\theequation}
\end{equation}
The loss function of $\alpha$ from \eqref{eq:alpha_loss} therefore becomes 
\begin{equation}
    \label{eq:alpha_loss_2}
    \mathcal{L}_\mathsf{LML}(\alpha,\beta^*(\alpha)) = - \sum_{m=1}^{M} \log p(\boldsymbol{y}^{(1:K)},(\boldsymbol{\theta}^{(1:K)}, \boldsymbol{\phi})^m_{\mathsf{t+1}}|\beta^*(\alpha)).
\end{equation}

An architecture sketch is illustrated in Figure \ref{fig:framework}. The complete algorithm, including posterior sampling and the bilevel optimization of closure and surrogate models, is provided in Algorithm \ref{alg:bilevel}. 

\afterpage{
  %\clearpage
\begin{algorithm}[H]
\caption{Joint Inverse Inference and Surrogate Training via Bilevel Optimization Framework}
\label{alg:bilevel}
\begin{algorithmic}[1]
\Require 
Observations $\mathbf{y}^{(1:K)}$; 
initial parameters $\alpha_0, \beta_0$;  
number of MALA chains $M$; 
initial MALA chains $X_0^{1:M}$;
number of upper-level iterations $\mathsf{T}$; 
number of lower-level iterations $N$;
MALA step size $\gamma$;
closure model step size $\eta_\alpha$;
surrogate model step size $\eta_\beta$;

\For{$\mathsf{t} = 0,1,\dots, \mathsf{T}-1$}

    \Statex \textbf{// Posterior sampling using ensemble MALA}
    \For{$m = 1,\dots,M$}
        \State Update chain $m$ via one MALA step: $X_{\mathsf{t}+1}^{m} = \mathrm{MALA}(X_\mathsf{t}^{m};\beta_{\mathsf{t} N},\gamma)$, where $X^{m}_\mathsf{t} = (\boldsymbol{\theta}^{(1:K)}, \boldsymbol{\phi})^{m}_\mathsf{t}$.
    \EndFor
    \Statex \textbf{// Lower-level: surrogate update}
    \State Select a random MALA chain index $p \sim \mathrm{Uniform}\{1,\dots,M\}$,
    \For{$n = 0,\dots,N-1$}    
        \State Compute $\mathcal{L}_\mathsf{Surrogate}(\alpha,\beta) = \mathcal{L}_{\text{FNO Super/FNO Phy/PINNs}} (\alpha,\beta)$, using $\left ( \boldsymbol{\theta}^{(1:K)} \right )^p_\mathsf{t+1}$.
        \State Update surrogate parameters: $\beta_{\mathsf{t} N+n+1} = \beta_{\mathsf{t} N+n} - \eta_\beta \nabla_\beta \mathcal{L}_\mathsf{Surrogate}(\alpha,\beta)$.
    \EndFor

    \Statex \textbf{// Upper-level: closure model update}
    \State Compute $\mathcal{L}_\mathsf{LML}(\alpha,\beta^*(\alpha)) = - \sum_{m=1}^{M} \log p(\boldsymbol{y}^{(1:K)}, (\boldsymbol{\theta}^{(1:K)}, \boldsymbol{\phi})^{m}_{\mathsf{t}+1} |\beta^*(\alpha))$, where $\beta^*(\alpha)=\beta_{(\mathsf{t}+1) N}$.
    \State Update closure parameters: $\alpha_{\mathsf{t}+1} = \alpha_\mathsf{t} - \eta_\alpha \nabla_\alpha \mathcal{L}_\mathsf{LML}(\alpha,\beta^*(\alpha))$.

\EndFor

\State \Return 
Posterior samples $X_\mathsf{T}^{(1:M)}$,
closure parameters $\alpha_\mathsf{T}$, 
surrogate parameters $\beta_{\mathsf{T} N}$.
\end{algorithmic}
\end{algorithm}
}

\subsubsection{Surrogate Model Training}\label{subsec5}

We consider two surrogate architectures in this work. The first one is FNO, which approximates mappings between function spaces by learning global convolution kernels in the Fourier domain. For each system, FNO takes as input the spatial and possibly temporal coordinates together with physical parameters $\boldsymbol{\theta}^{(k)}$, and outputs the whole predicted solution field $\hat{\mathbf{u}}^{(k)}$. We train FNO using two types of loss functions. The first one is supervised loss, where $F^{\beta}(\boldsymbol{\theta}^{(k)})$ is matched to states computed from the numerical solver, and the loss function is equal to
\begin{equation}
    \label{eq:supervised_loss}
    \mathcal{L}_{\text{FNO Super}}(\alpha,\beta) = \frac{1}{K} \sum_{k=1}^{K}  \left \| F^{\beta}(\boldsymbol{\theta}^{(k)}) - F^\dagger(\boldsymbol{\theta}^{(k)},\alpha) \right \| ^2,
\end{equation}
with $\boldsymbol{\theta}^{(k)}$ being an element from the selected MALA chain $(\boldsymbol{\theta}^{(1:K)}, \boldsymbol{\phi})^p_\mathsf{t+1}$.
The second method is purely physics-based loss, where finite differences are used to compute ODE/PDE residual, enforcing physical consistency with the governing equations. The loss function is represented by
\begin{align}
    \label{eq:fno_physics_loss}
    \mathcal{L}_{\text{FNO Phy}}(\alpha,\beta) &= \frac{1}{K} \sum_{k=1}^{K} \left \| \mathcal{R}(\boldsymbol{\theta}^{(k)},\alpha,\beta)\right \| ^2 ,
    % &= \left \|\mathcal{G}(u_k,\mathcal{N}_{\alpha}(u_k),x,t,\theta_k)-s(x,t)\right \| ^2,
\end{align}
where $\mathcal{R}$ calculates the physical residual using finite differences on the output of the surrogate $F^{\beta}(\boldsymbol{\theta}^{(k)})$. In certain PDE settings, weak-form residual is employed as well, which will be discussed in detail in Section \ref{sec5}.

We also employ PINNs as an alternative surrogate architecture. Unlike standard PINNs, which take only spatial-temporal coordinates as input, our formulation augments the input with physical parameters $\boldsymbol{\theta}^{(k)}$. This enables PINNs to represent an entire family of dynamical systems rather than a single instance, thus approximate the mapping from $\boldsymbol{\theta}^{(k)}$ to $\mathbf{u}^{(k)}$ for multiple systems. Training is carried out using a purely physics-informed objective and the loss function follows the structure of conventional PINNs, consisting of physics residuals of governing ODEs/PDEs evaluated at collocations points, together with penalty terms enforcing initial and boundary conditions. The loss function is
\begin{equation}
    \label{eq:PINNs_loss}
    \mathcal{L}_{\text{PINNs}}(\alpha,\beta) = \frac{1}{K} \sum_{k=1}^{K} \left (\mathcal{L}_{\text{Physics}}(\boldsymbol{\theta}^{(k)},\alpha) +\mathcal{L}_{\text{BC/IC}}(\boldsymbol{\theta}^{(k)}) \right), 
\end{equation}
where $\mathcal{L}_{\text{Physics}}$ is calculated based on provided equations and computed by automatic differentiation. And
\begin{equation}
    \label{eq:BC_loss}
    \mathcal{L}_{\text{BC/IC}}(\boldsymbol{\theta}^{(k)}) =  \| F^{\beta}|_{\partial \Omega}\,(\boldsymbol{\theta}^{(k)}) - \mathbf{g}_{\text{BC}}^{(k)} \| ^2 + \| F^{\beta}|_{t=0}(\boldsymbol{\theta}^{(k)}) - \mathbf{g}_{\text{IC}}^{(k)} \|^2  ,
\end{equation}
where $\mathbf{g}_{\text{BC}}^{(k)}$ and $\mathbf{g}_{\text{IC}}^{(k)}$ denote boundary and initial conditions.

Together, FNO-based and PINN-based surrogates enable efficient approximation of the forward operator $F^\dagger$ across multiple systems, while providing the differentiability required for gradient-based sampling within the ensemble MALA framework. Crucially, it is important to note that these surrogate models are not pre-trained offline. Instead, they are optimized jointly with the inverse problem in an online manner. This coupled training strategy ensures that the surrogates are adaptively refined in the parameter regions explored by the MALA sampler.

\section{Numerical Experiments}\label{sec5}

In this section, we validate our proposed methodology on three types of differential equations: a nonlinear mass-damper ODE, a nonlinear 2D Darcy flow and a generalized Burgers' equation, corresponding to contribution \ref{con:c4}.

\subsection{Nonlinear Mass-Damper System}\label{subsec6}

Our first experiment investigates a nonlinear mass-damper system. In this setup, the stiffness, initial position and velocity of each system are treated as unknown physical parameters, while the nonlinear damping law serves as the unknown closure model.

\subsubsection{Problem Formulation}

\begin{figure}[!t]
\centering
\includegraphics[scale=0.4]{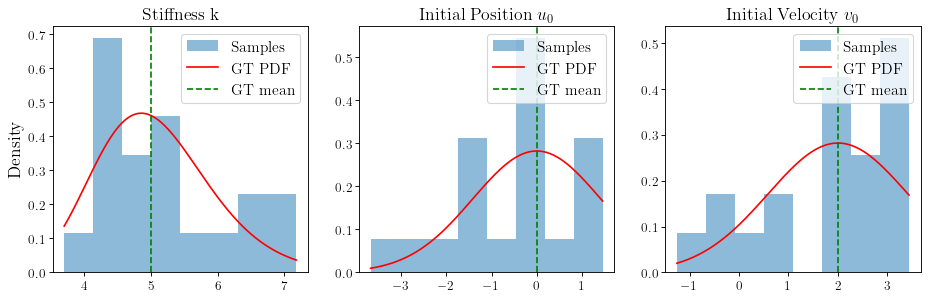}
\caption{\textbf{Distribution of ground-truth physical parameters for $\mathbf{K=20}$ mass–damper systems in Experiment 1.} Each system’s parameter vector $\boldsymbol{\theta} = \{\log(k), u_0, v_0\}$ is sampled from the hierarchical prior with hyperparameters $\boldsymbol{\mu}_\phi = \{\log(5.0), 0.0, 2.0\}$ and $\boldsymbol{\tau}_\phi = \{0.03, 2.0, 2.0\}$. The histograms show the parameter distributions across the 20 systems, while the overlaid curves represent the corresponding ground-truth probability density functions implied by the hierarchical prior. The green dashed lines indicate ground-truth population means. The apparent deviation of the histograms from the ground-truth curves occurs because a limited population of only 20 systems poorly represents the full underlying probability distribution. } 
\label{GT_parameters_1}
\end{figure}

The nonlinear mass-damper system is defined by
\begin{equation}
\label{eq:mass_spring_1}
\begin{aligned}
\ddot{u}(t) + f(\dot{u}(t)) + k\, u(t) &= 10\sin(t), \quad t \in [0,8],\\
u(0) &= u_0,\\
\dot{u}(0) &= v_0.
\end{aligned}
\end{equation}
where $u$ is the displacement of mass and the nonlinear damping $f(\dot{u})=0.08\dot{u}^3+0.08\dot{u}$ is shared among all systems. The stiffness k, initial position and velocity $u_0,v_0$ vary for each system. To ensure the stiffness $k$ is always positive, parameters to be inferred are set to $\boldsymbol{\theta}=\{ \log(k), u_0, v_0\}$. The ground truth hyperparameters for the population are $\boldsymbol{\mu}_{\phi}=\{ \log(5.0),0.0,2.0\}$ and $\boldsymbol{\tau}_\phi=\{ 0.03,2.0,2.0\}$. Following \eqref{hierarchical}, ground truth parameters $\boldsymbol{\theta}$ are generated from the hyperprior, with histograms of parameters of $K=20$ generated systems shown in Figure \ref{GT_parameters_1}.

\subsubsection{Solver}

\begin{figure}[!t]
\centering
\includegraphics[scale=0.35]{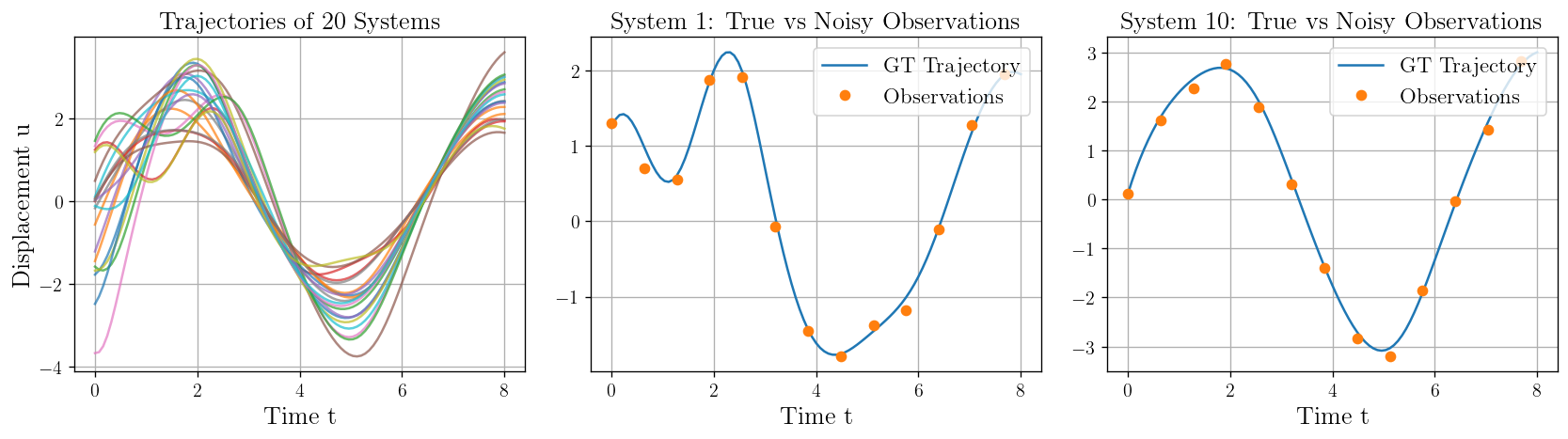}
\caption{\textbf{Displacement trajectories $\mathbf{u(t)}$ for 20 nonlinear mass–damper systems.} These systems are governed by \eqref{eq:mass_spring_1} and solved using the leapfrog integrator with system-specific parameters in Figure~\ref{GT_parameters_1}. The right two panels illustrate, for two representative systems, the corresponding sparse and noisy observations generated via observation operator $g^{(k)}$.  }
\label{GT_trajectories_1}
\end{figure}

The numerical solver used to solve \eqref{eq:mass_spring_1} is the leapfrog integrator, a second-order scheme commonly employed for oscillatory systems. Using the leapfrog solver with timestep $\Delta t = 0.08s$, Figure \ref{GT_trajectories_1} displays the displacement trajectories generated for the $K=20$ systems corresponding to the parameter values shown in Figure~\ref{GT_parameters_1}. In addition, two representative systems are further shown with their corresponding sparse and noisy measurements $\mathbf{y}$, which are obtained every 8 steps. In the first experiment, the observation operator $g^{(k)}$ is the same across systems and the observation noise is set to $\sigma = 0.15$.
 
\subsubsection{Surrogate Implementation Details}

We evaluate four models for the mass–damper system. The first baseline solves the inverse problem directly using the leapfrog integrator, without introducing any surrogate model, as illustrated in Section \ref{subsec3}. To avoid the use of iterative solvers during Bayesian inference, we then construct two different surrogates, FNOs and PINNs, training them in a bilevel scheme together with the inverse problem. 

As mentioned in Section \ref{subsec5}, the FNO surrogate is trained using two alternative loss formulations: a purely physics-based residual loss and a supervised loss. In the mass–damper setting, the FNO receives as input the parameter vector $\boldsymbol{\theta}^{(k)}$ and time coordinates $\mathbf{t}$, and outputs both displacement $\boldsymbol{\mathsf{w}}$ and velocity $\boldsymbol{\mathsf{v}}$.
% i.e., $F^{\beta}:\boldsymbol{\theta} \mapsto (\mathsf{w}(t),\mathsf{v}(t))$. 
Producing both quantities enables more accurate temporal derivatives via finite differences. The physics-based loss $\mathcal{L}_{\text{FNO Phy}}$ in \eqref{eq:fno_physics_loss} is the summation of two residual terms:
\begin{align}
    \label{eq:mass_damper_residual}
    \mathcal{R} &= \mathcal{R}_1 + \mathcal{R}_2, \nonumber \\ 
    \mathcal{R}_1 &= \left \| \dot{\boldsymbol{\mathsf{w}}} - \boldsymbol{\mathsf{v}} \right \| ^2, \\
    \mathcal{R}_2 &= \left \| \dot{\boldsymbol{\mathsf{v}}} - (10\sin (\mathbf{t}) - f^{\alpha}(\boldsymbol{\mathsf{v}})-k\boldsymbol{\mathsf{w}}) \right \| ^2, \nonumber
\end{align}
where the time derivatives $\dot{\boldsymbol{\mathsf{w}}}$ and $\dot{\boldsymbol{\mathsf{v}}}$ are approximated using centered finite differences. The supervised FNO loss $\mathcal{L}_{\text{FNO Super}}$, defined in \eqref{eq:supervised_loss}, penalizes discrepancies between FNO outputs and reference trajectories computed by the leapfrog integrator.

The PINN surrogate takes as inputs $\boldsymbol{\theta}^{(k)}$ and $t$ as well and predicts only the displacement. The required first and second-order derivatives are obtained through automatic differentiation, enabling PINN to enforce the governing equation in \eqref{eq:mass_spring_1}. More implementation details regarding network structures and training are provided in Appendix \ref{secA}.
\begin{figure}[!t]
\centering
\includegraphics[scale=0.29]{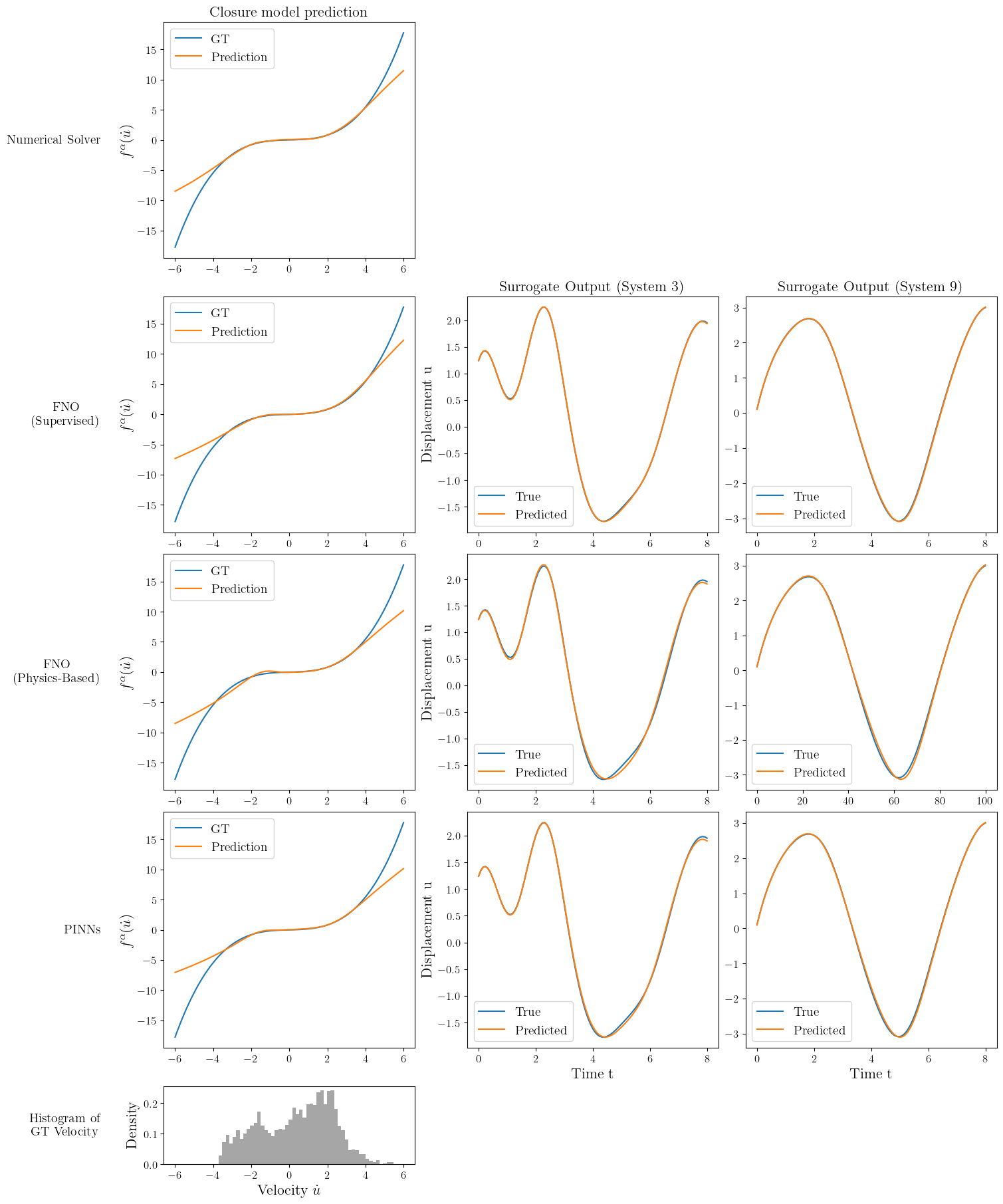}
\caption{ \textbf{Comparison of closure model estimation and surrogate performance across different models with $\mathbf{K=20}$ in Experiment 1.} The rows correspond to Numerical Solver, Supervised FNO, Physics-Based FNO, and PINNs, respectively. \textbf{Left Column:} Comparison between the learned closure $f^\alpha(\dot{u})$ (orange) and the ground truth $f(\dot{u})$ (blue) over the velocity range $\dot{u} \in [-6, 6]$. \textbf{Middle $\&$ Right Columns:} Surrogate predictions of displacement $u(t)$ for two selected systems (System 3 and System 9). \textbf{Bottom Panel:} Histogram of velocity values in GT dataset. }
\label{alpha_and_beta_1}
\end{figure}

In the following section, we evaluate the performance of the four models considered in this work: the baseline approach that performs \textbf{inverse problem using a numerical forward solver}, and three surrogate-accelerated approaches in which the forward solver is replaced by an \textbf{FNO trained either with supervised or physics-based losses}, or by a \textbf{PINN-based surrogate}.

\subsubsection{Results}

We structure our evaluation into three main parts: a performance comparison of the four models, a comparison between the hierarchical and non-hierarchical Bayesian settings, and an analysis of computational efficiency.

\vspace{1ex}
\noindent\textbf{Performance Comparison on Inverse and Forward Problems.} 

% \paragraph{Model Performance on Inverse and 
% Forward Problems}
We compare the four models with respect to three key criteria: (i) accuracy in inferring physical parameters, (ii) accuracy in estimating the nonlinear closure model, and (iii) quality of the surrogate approximation to the forward dynamics. Experiments are conducted across various values of $K$, ranging from $5$ to $100$. However, the method utilizing a numerical solver as the forward mapping is not implemented for cases where $K > 30$. This limitation is due to memory overflow, as the multiple temporal iterations of the solver cause excessive memory consumption when computing gradients as required by MALA.

\begin{figure}[!t]
\centering
\includegraphics[scale=0.41]{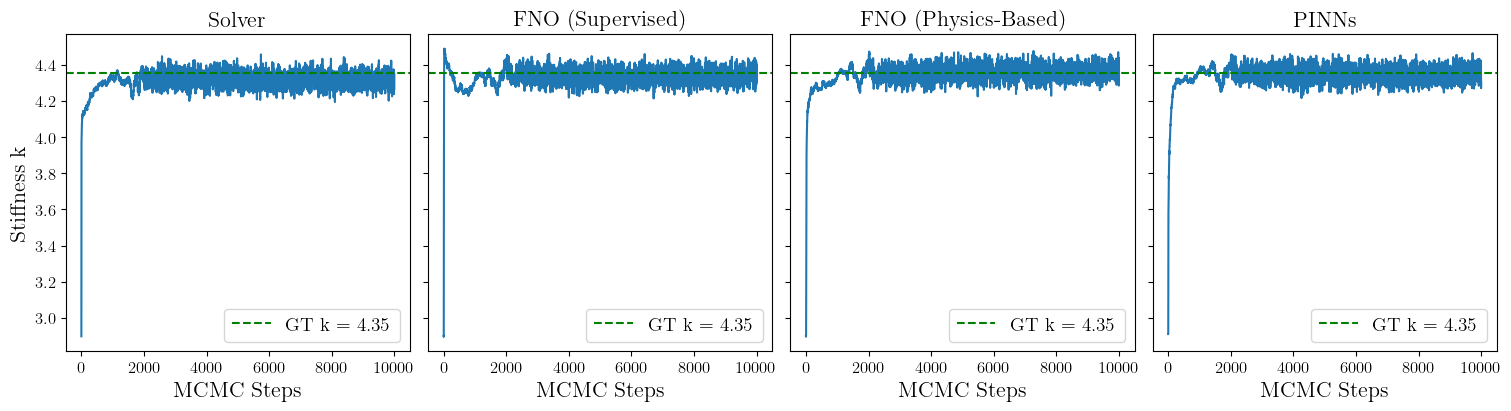}
\caption{\textbf{Trace plots illustrating the evolution of a randomly selected MALA chain for stiffness $k$ of System 9 across four models with $\mathbf{K=20}$ in Experiment 1.} The figures illustrate convergence behaviour relative to the GT parameter value (green dashed line). }
\label{four_model_mcmc_1}
\end{figure}
\begin{table}[!t]
\centering
\caption{\textbf{Quantitative comparison of the four models across varying numbers of systems ($\mathbf{K \in \{5, 10, 20, 30, 50, 100\}}$) in Experiment 1.} The table evaluates \textbf{Parameter Inference} performance via the MSE of posterior means relative to GT parameter values and \textbf{Coverage}, defined as the percentage of GT parameters falling within two posterior standard deviations. Additionally, \textbf{Closure MSE} measures the MSE of the learned nonlinear closure $f^\alpha(\dot{u})$ evaluated on $\dot{u} \in [-5,5]$, while \textbf{Surrogate MSE} quantifies the accuracy of the surrogate model's trajectory predictions compared to GT trajectories.}
\label{tab:result1_models_performance}
\renewcommand{\arraystretch}{1.3}

\resizebox{0.55\linewidth}{!}{%
\begin{tabular}{cccccc}
\toprule
\textbf{Model} 
& \makecell{\textbf{No. of}\\\textbf{Systems}}
& \multicolumn{2}{c}{\textbf{Parameter Inference}} 
& \textbf{Closure MSE} 
& \textbf{Surrogate MSE} 
\\
\cmidrule(lr){3-4}
& & \textbf{Mean MSE} & \textbf{Coverage} & & \\ 
\midrule

\multirow{4}{*}{\textbf{Solver}}

 & $K = 5$ & $1.44 \times 10^{-2}$ & $100 \%$ & $1.05$ & - \\
 & $K = 10$ & $2.41 \times 10^{-2}$ & $93.3 \%$ & $0.48$ & - \\
 & $K = 20$ & $2.70 \times 10 ^{-2}$ & $96.7 \%$ & $0.50$ & - \\
 & $K = 30$  & $3.02 \times 10^{-2}$ & $91.1 \%$ & $0.45$ & - \\

\midrule

\multirow{6}{*}{\makecell{\textbf{FNO}\\\textbf{(Supervised)}}}
 & $K = 5$ & $4.85 \times 10^{-2}$ & $93.3 \%$ & $1.05$ & $1.53 \times 10^{-3}$ \\
 & $K = 10$ & $1.74 \times 10^{-2}$ & $96.7 \%$ & $0.84$ & $8.60 \times 10^{-4}$ \\
 & $K = 20$ & $3.13 \times 10^{-2}$ & $93.3\%$ & $0.88$ & $8.44 \times 10^{-4} $  \\
 & $K = 30$  & $3.42 \times 10^{-2}$ & $93.3\%$ & $0.88$ & $1.24 \times 10^{-3}$ \\
 & $K = 50$  & $3.56 \times 10^{-2}$ & $85.3 \%$ & $0.58$ & $9.37 \times 10^{-4}$ \\
 & $K = 100$ & $4.75 \times 10^{-2}$ & $79.7 \%$ & $0.69$ & $9.68 \times 10^{-4}$ \\
\midrule

\multirow{6}{*}{\makecell{\textbf{FNO}\\\textbf{(Physics-based)}}}
 & $K = 5$ & $1.51 \times 10^{-1}$ & $73.3 \%$ & $4.08$ & $1.43 \times 10^{-2}$ \\
 & $K = 10$ & $1.27 \times 10^{-1}$ & $60.0 \%$ & $2.16$ & $1.62 \times 10 ^{-2}$ \\
 & $K = 20$ & $6.77 \times 10 ^{-2}$ & $86.7\%$ & $1.38$ & $4.58 \times 10^{-3}$ \\
 & $K = 30$  & $2.62 \times 10^{-1}$ & $73.3 \%$ & $1.54$ & $1.44 \times 10^{-2}$ \\
 & $K = 50$  & $6.86 \times 10^{-2}$ & $78.7 \%$ & $1.30$ & $6.31 \times 10^{-3}$ \\
 & $K = 100$ & $7.38 \times 10^{-2}$ & $63.3 \%$ & $1.26$ & $7.78 \times 10^{-3}$ \\
\midrule

\multirow{6}{*}{\textbf{PINNs}}
 & $K = 5$ & $4.19 \times 10^{-2}$ & $86.7 \%$ & $1.18$ & $3.80 \times 10^{-3}$\\
 & $K = 10$ & $2.44 \times 10^{-2}$ & $93.3 \%$ & $1.15$ & $1.66 \times 10^{-3}$ \\
 & $K = 20$ & $6.37 \times 10^{-2}$ & $90.0\%$ & $1.27$ & $3.97 \times 10 ^{-3}$ \\
 & $K = 30$  & $5.59 \times 10^{-2}$ & $92.2 \%$ & $1.15$ & $2.64 \times 10^{-3}$ \\
 & $K = 50$  & $5.06 \times 10^{-2}$ & $81.3 \%$& $0.79$ & $2.29 \times 10^{-3}$\\
 & $K = 100$ & $5.77 \times 10^{-2}$ & $70.0 \%$ & $1.02$ & $3.30 \times 10^{-3}$ \\
\bottomrule

\end{tabular}%
}
\end{table}

Figure \ref{alpha_and_beta_1} summarizes the performance of four models in estimating the closure model and in predicting system trajectories using the learned surrogate, with $K$ setting to 20. Since the closure $f$ in this experiment is evaluated on the velocity $\dot{u}$, we additionally plot the distribution of velocities extracted from ground-truth trajectories. This distribution reveals the range of velocities that exists in ground-truth data, and correspondingly explains why the learned closure model exhibits higher accuracy in this region. Quantitative results, including the mean squared error (MSE) of the closure estimation (evaluated over the velocity range $\dot{u} \in [-5, 5]$, as the vast majority of ground-truth velocity data falls within this interval) and the MSE of surrogate model predictions across all systems, are reported in Table \ref{tab:result1_models_performance}. 

\begin{figure}[!t]
\centering
\includegraphics[scale=0.28]{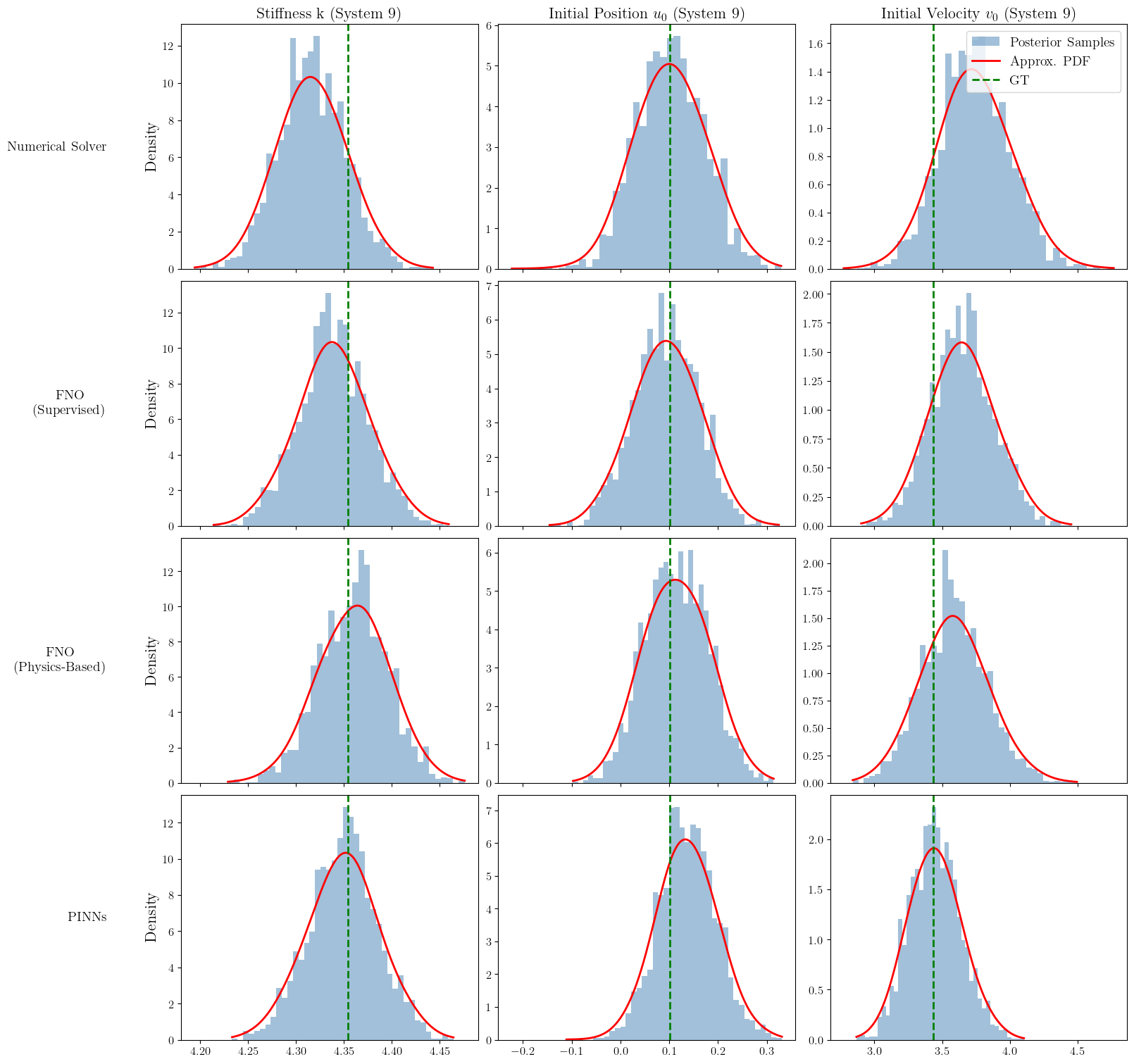}
\caption{\textbf{Posterior histograms and corresponding approximated probability density functions for three parameters from System 9 across four models with $\mathbf{K=20}$ in Experiment 1.} Each panel shows the marginal posterior distribution obtained from the final 1,000 samples of one random MALA chain after convergence. The blue bars represent the posterior samples, the red curve denotes the approximated density, and the green dashed line marks the GT parameter value. }
\label{four_model_histograms_1}
\end{figure}

According to Figure \ref{alpha_and_beta_1} and Table \ref{tab:result1_models_performance}, the numerical solver and supervised FNO demonstrate high accuracy in closure model prediction, effectively capturing the cubic nonlinearity. PINNs also yield relatively accurate results. In contrast, the physics-based FNO performs poorly, failing to provide accurate predictions with significantly large errors observed at $K=5,10,30$ as shown in Table \ref{tab:result1_models_performance}. Furthermore, Figure \ref{alpha_and_beta_1} illustrates that the accuracy of the learned closure model is dependent on data availability, exhibiting superior performance in ranges with a high density of training data. Regarding surrogate model training, the supervised FNO demonstrates high accuracy, consistently achieving lowest MSE across all settings. PINNs rank second, delivering competitive performance with errors in the $10^{-3}$ range, slightly higher than supervised FNO. In contrast, the physics-based FNO is the least accurate, with errors deteriorating significantly to the $10^{-2}$ range, approximately one order of magnitude higher than the other methods.

Next, we assess the performance of four models on parameter inference. Figure \ref{four_model_mcmc_1} illustrates the evolution of a random chain chosen from $M$ MALA chains for stiffness $k$ in System 9, showing how the chains produced by each model converge towards their respective posterior distributions. As mentioned in Section \ref{subsec3}, the sampling process comprises two stages: the initial 9,000 samples were generated during the joint training of inverse and forward problems, followed by an additional 1,000 samples collected via ensemble MALA with fixed $\alpha$, $\beta$ and preconditioner $\mathbf{C}$. Figure \ref{four_model_histograms_1} further presents, for each model, the posterior histograms for the final 1,000 steps together with the corresponding approximated probability density functions for all three parameters of System 9. A quantitative comparison on parameter inference across the four models is summarized in Table \ref{tab:result1_models_performance} as well. The table reports the MSE between estimated posterior means and ground-truth parameters, averaged over all systems and all inferred parameters. The coverage of estimated posterior distributions is also reported, measured as the percentage of ground-truth parameter values lying within two posterior standard deviation. These statistics jointly assess the accuracy of inferred parameters and the quality of posterior uncertainty quantification. In addition, during our experiments, we noted that omitting either the bias-correcting Metropolis step or the ensemble-based preconditioner leads to non-convergent Langevin dynamics, demonstrating the essential role these components play in obtaining a robust and stable algorithm.
% mechanisms play in achieving accurate and stable posterior distributions.

Overall, the numerical solver and supervised FNO deliver the most robust parameter inference, achieving the lowest MSE and consistently high coverage rates. PINNs demonstrate competitive performance, establishing itself as a reliable physics-based alternative with only marginally lower accuracy. Conversely, physics-based FNO suffers from poorly calibrated posterior distributions, its substantially higher MSE and low coverage (dropping to 60$\%$ at $K=10$) indicate significant estimation error, where ground-truth values frequently fall outside predicted posterior distributions.

Furthermore, the comparison across varying numbers of systems $K$ in Table \ref{tab:result1_models_performance} reveals a distinct trend. As the number of available systems increases, the accuracy of both the closure model and the surrogate prediction generally improves, likely due to the larger amounts of training data. Conversely, the accuracy of parameter inference tends to degrade. This decline is attributed to the inherent difficulty of performing inference on a higher-dimensional posterior distribution.

\begin{table}[h]
\centering
\caption{\textbf{Comparison of hierarchical and non-hierarchical frameworks trained using PINNs as surrogate model in Experiment 1.} Reported metrics are the same as in Table \ref{tab:result1_models_performance}.}
\renewcommand{\arraystretch}{1.3}
\resizebox{0.6\linewidth}{!}{%
\begin{tabular}{ccccccc}
\toprule
\textbf{Method} 
\label{tab:hierarchical_comparison_1}
& \makecell{\textbf{No. of}\\\textbf{Systems}}
& \multicolumn{2}{c}{\textbf{Parameter Inference}}
& \textbf{Closure MSE}
& \textbf{Surrogate MSE}
\\
\cmidrule(lr){3-4}
& & \textbf{Mean MSE} & \textbf{Coverage} & & \\
\midrule

\multirow{5}{*}{\textbf{Hierarchical}}
 & $K = 5$ & $4.19 \times 10^{-2}$ & $86.7 \%$ & $1.18$ & $3.80 \times 10^{-3}$\\
 & $K = 10$ & $2.44 \times 10^{-2}$ & $93.3 \%$ & $1.15$ & $1.66 \times 10^{-3}$ \\
 & $K = 20$ & $6.37 \times 10^{-2}$ & $90.0\%$ & $1.27$ & $3.97 \times 10^{-3}$ \\
 & $K = 30$  & $5.59 \times 10^{-2}$ & $92.2 \%$ & $1.15$ & $2.64 \times 10^{-3}$ \\
 & $K = 50$  & $5.06 \times 10^{-2}$ & $81.3 \%$& $0.79$ & $2.29 \times 10^{-3}$\\
 & $K = 100$ & $5.77 \times 10^{-2}$ & $70.0 \%$ & $1.02$ & $3.30 \times 10^{-3}$ \\
\midrule

\multirow{5}{*}{\makecell{\textbf{Non-}\\\textbf{Hierarchical}}}
 & $K = 5$ &$4.79 \times 10^{-1}$ & $26.7 \%$ & $36.89$ & $2.25 \times 10^{-1}$ \\
 & $K = 10$ & $2.55 \times 10^{-2}$ & $83.3 \%$ & $1.52$ & $2.42 \times 10^{-3}$ \\
 & $K = 20$ & $6.02 \times 10 ^{-2}$ & $86.7 \%$ & $1.44$ & $5.41\times10^{-3}$ \\
 & $K = 30$  & $9.56 \times 10^{-2}$ & $73.3 \%$ & $1.25$ & $9.76 \times 10^{-3}$ \\
 & $K = 50$  & $6.74 \times 10^{-2}$ & $70.0 \%$ & $0.99$ & $3.68 \times 10^{-3}$\\
 & $K = 100$ & $1.01 \times 10^{-1}$ & $55.7 \%$ & $1.51$ & $5.13 \times 10^{-3}$ \\
% \midrule

% \multirow{3}{*}{\textbf{Deterministic}}
%  & $n = 5$  & -- & -- & -- & -- \\
%  & $n = 10$ & -- & -- & -- & -- \\
%  & $n = 20$ & -- & -- & -- & -- \\
\bottomrule
\end{tabular}%
}
\end{table}
\begin{figure}[!t]
\centering
\includegraphics[scale=0.35]{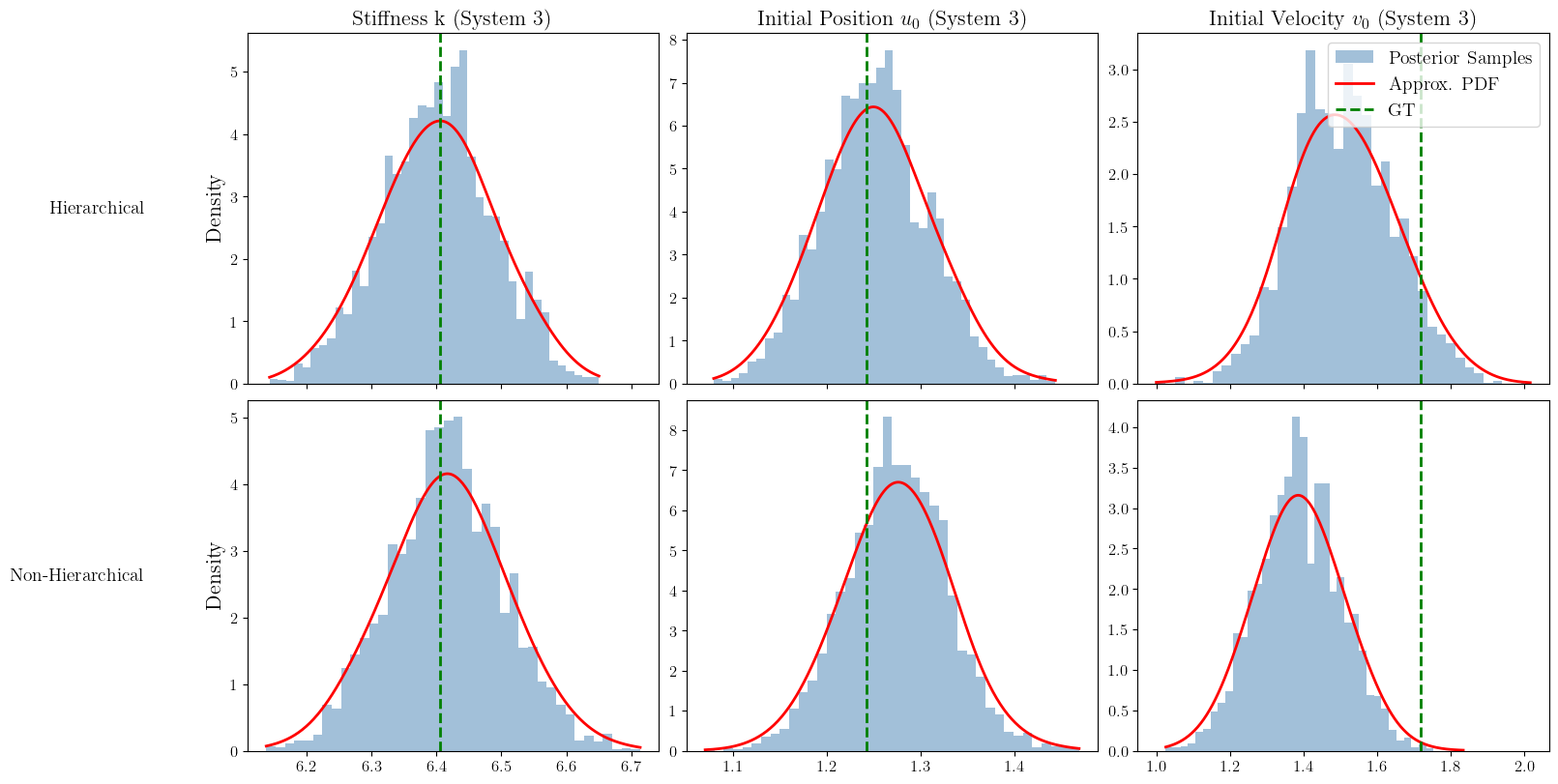}
\caption{\textbf{Comparison of posterior parameter distributions between Hierarchical and Non-Hierarchical Bayesian frameworks using a PINN-based surrogate with $\mathbf{K=20}$ in Experiment 1. } The figure compares the uncertainty quantification performance for three parameters from System 3. The \textbf{top row (Hierarchical)} and \textbf{bottom row (Non-Hierarchical)} show distinct posterior densities. GT values are indicated by green dashed lines, overlaid on the posterior samples (blue histograms) and their PDF approximations (red curves). }
\label{hierarchical_histograms_1}
\end{figure}

\vspace{1ex}
\noindent\textbf{Hierarchical and Non-Hierarchical Bayes}

We now perform an ablation study, where we explore the impact of the hierarchical Bayesian formulation by comparing it against the non-hierarchical alternative. In the non-hierarchical setting, each parameter is treated independently, and no population-level hyperparameters $\boldsymbol{\phi}$ are inferred. We test both formulations using PINNs as the surrogate forward model, trained jointly with the inverse problem within the bilevel framework. Table \ref{tab:hierarchical_comparison_1} reports the performance of these two approaches using the same metrics as in Table \ref{tab:result1_models_performance}.

\begin{figure}[!t]
\centering
\includegraphics[scale=0.22]{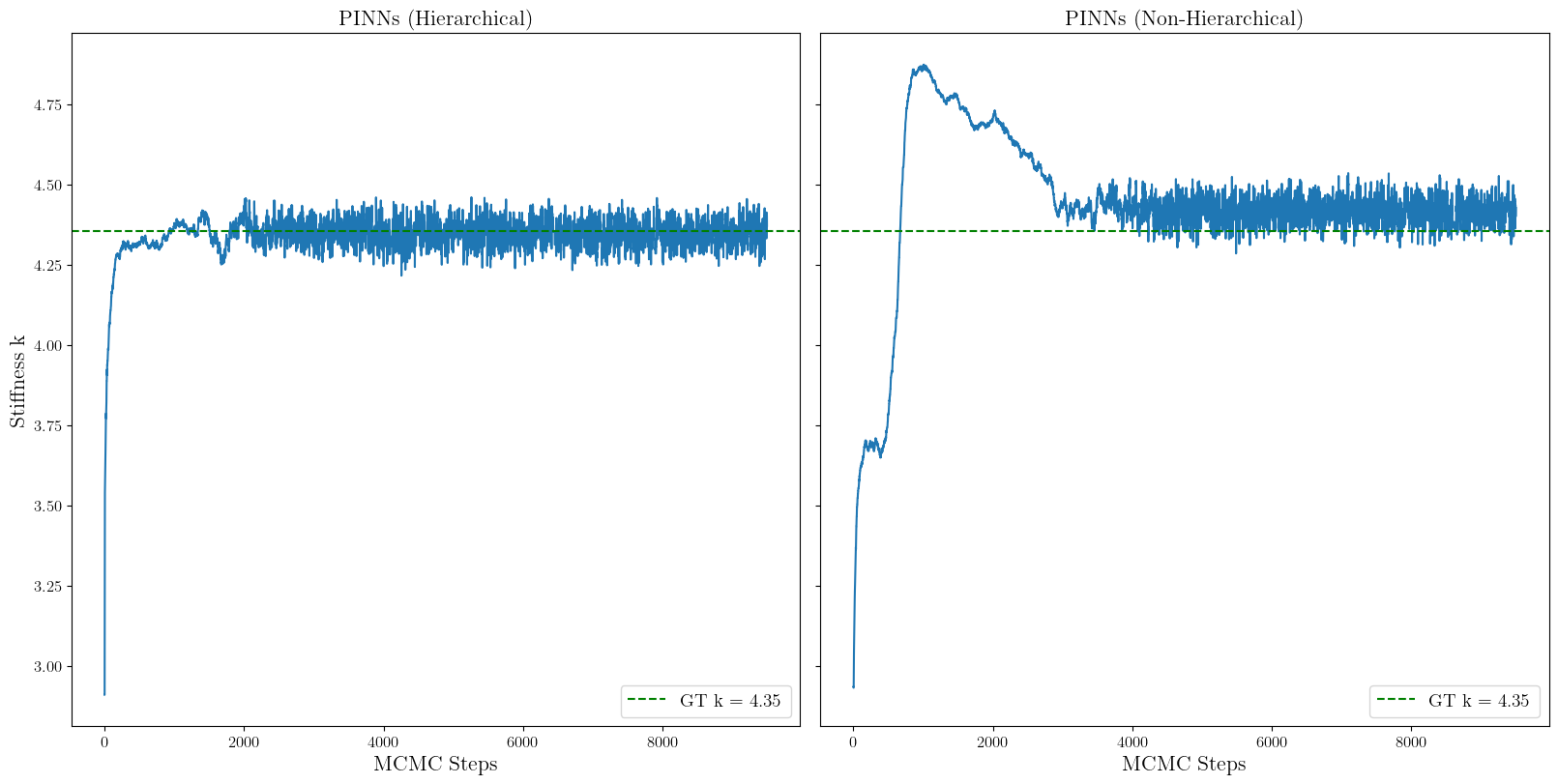}
\caption{\textbf{MALA chains of posterior samples for stiffness $k$ of System 9 for Hierarchical and Non-Hierarchical Bayesian frameworks in Experiment 1.} The figures compare the convergence rate between hierarchical and non-hierarchical models. }
\label{MCMC_hie_1}
\end{figure}

A fundamental mechanism of hierarchical Bayes is its ability to borrow strength from the population, which acts to produce accurate parameter estimates and reduce estimation uncertainty. For posterior histograms shown in Figure \ref{hierarchical_histograms_1}, the visual reduction in posterior standard deviation for the hierarchical model appears modest, with estimation uncertainties closely resembling those of the non-hierarchical alternative. However, quantitative results in Table \ref{tab:hierarchical_comparison_1} reveal a significant advantage in accuracy of the hierarchical setting. Furthermore, observations during training indicate that the non-hierarchical model converges at a noticeably slower rate than the hierarchical approach, which can be seen from MALA trace plots in Figure \ref{MCMC_hie_1}.

\begin{figure}[!t]
\centering
\includegraphics[scale=0.38]{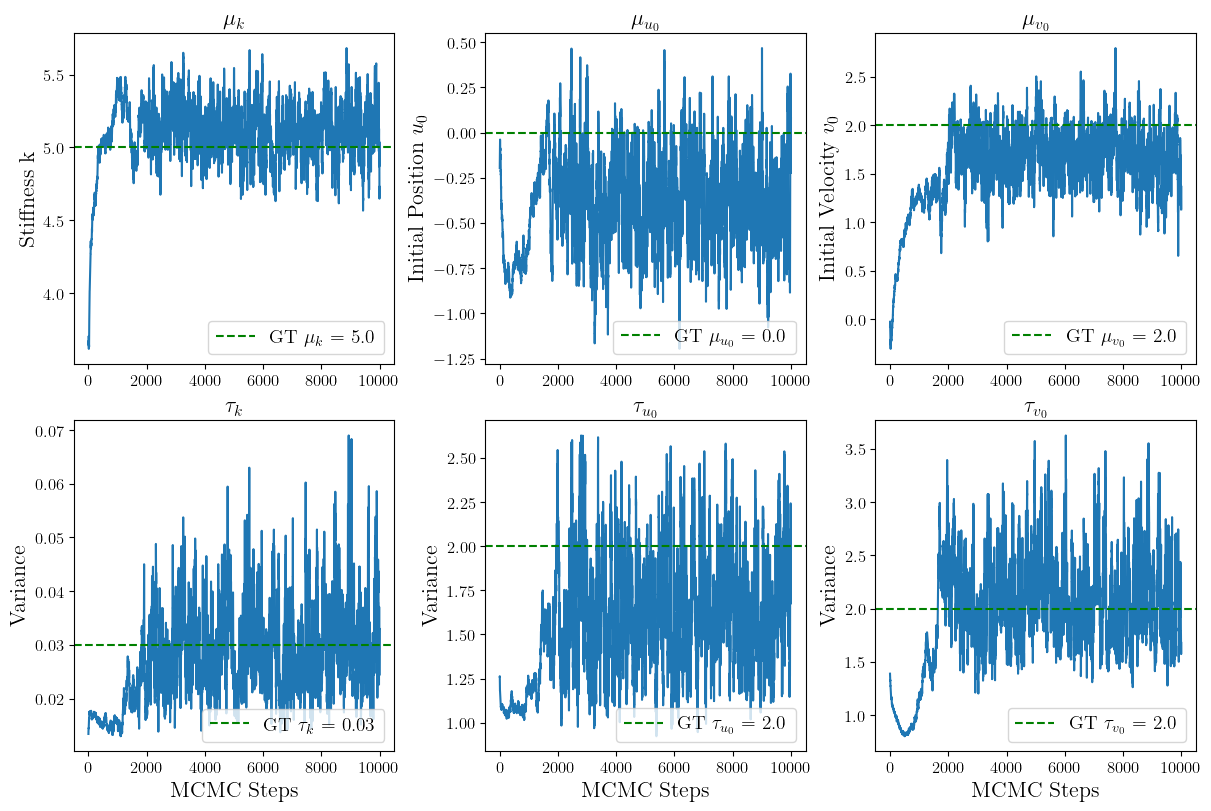}
\caption{\textbf{MALA chains of posterior samples for hyperparameters inferred using the PINN-based hierarchical Bayesian framework in Experiment 1.} The panels display the MALA chains for $\boldsymbol{\mu}_\phi$ and $\boldsymbol{\tau}_\phi$. The top row corresponds hyperprior means ($\mu_k, \mu_{u_0}, \mu_{v_0}$), while the bottom row shows the variances ($\tau_k, \tau_{u_0}, \tau_{v_0}$). Green dashed lines represent ground-truth values.}
\label{hyperprior_MCMC_1}
\end{figure}
\begin{table}[!t]
\centering
\caption{\textbf{Computational efficiency comparison across considered models for varying number of systems ($\mathbf{K \in \{5, 10, 20, 30, 50, 100\}}$) in Experiment 1.} The reported values represent the average time (in seconds) required to complete one full training epoch within the bilevel framework.}
\resizebox{0.8\linewidth}{!}{%
\begin{tabular*}{\textwidth}{@{\extracolsep{\fill}}cccccc}
\toprule
\textbf{Samples} & \textbf{Solver} & \makecell{\textbf{FNO}\\\textbf{(Supervised)}} & \makecell{\textbf{FNO}\\\textbf{(Physics-based)}} & \textbf{PINNs} & \makecell{\textbf{PINNs}\\\textbf{(Non-Hierarchical)}} \\
\midrule
$K=5$  & 0.140s & 0.549s & 0.099s & 0.038s & 0.040s \\
$K=10$ & 0.171s & 0.615s & 0.104s & 0.038s & 0.042s \\
$K=20$ & 0.201s & 0.712s & 0.124s & 0.040s & 0.045s \\
$K=30$ & 0.238s & 0.763s & 0.157s & 0.041s & 0.046s \\
$K=50$ & - & 0.842s & 0.232s & 0.048s & 0.058s \\
$K=100$ & - & 1.158s & 0.489s & 0.091s & 0.091s \\
\bottomrule
\label{tab:result1_efficiency}
\end{tabular*}%
}
\end{table}

In addition, the hierarchical formulation yields estimates of population-level hyperparameters $\boldsymbol{\phi}$, which are not available in the non-hierarchical setting. Figure \ref{hyperprior_MCMC_1} shows the MALA traces of inferred hyperprior components. The mean values of inferred hyperpriors are $\overline{\boldsymbol{\mu}}_\phi=\{ 5.08, -0.36, 1.70\}$ and $\overline{\boldsymbol{\tau}}_\phi=\{ 0.030, 1.64, 2.08\}$, compared to ground-truth values $\boldsymbol{\mu}_\phi=\{ 5.0, 0.0, 2.0\}$ and $\boldsymbol{\tau}_\phi=\{ 0.03,2.0,2.0\}$. Crucially, these learned population statistics can serve as informative, data-driven priors for future inference on new systems, thereby eliminating the reliance on arbitrary prior selection and enhancing the overall inference robustness.

\vspace{1ex}
\noindent\textbf{Efficiency of different algorithms}

Finally, we evaluate the computational efficiency of different frameworks. Table \ref{tab:result1_efficiency} reports the average time required to complete one full training epoch for all models. For the baseline model using numerical solver as forward mapping, the reported computational time includes the time for updating closure model parameters and the ensemble MALA step, while for other methods, the recorded time includes the whole bilevel iteration, including the closure update, the MALA inference step, and the training step of surrogate model. 

\begin{figure}[!t]
\centering
\includegraphics[scale=0.38]{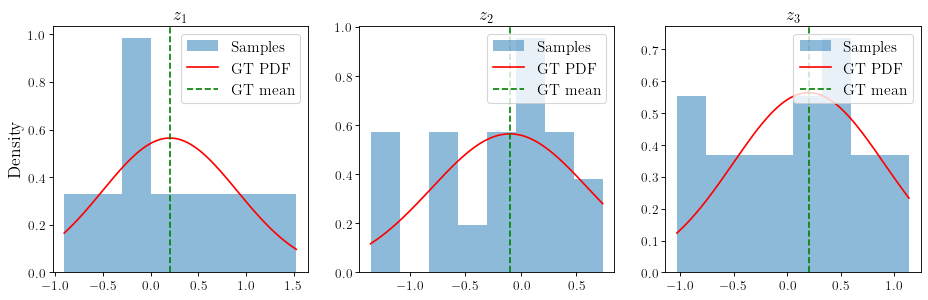}
\caption{\textbf{Distribution of ground-truth physical parameters for 20 Darcy flow systems in Experiment 2.} Each system’s parameter vector $\boldsymbol{\theta} = \{z_1,z_2,z_3\}$ is sampled from the hierarchical prior with hyperparameters $\boldsymbol{\mu}_\phi = \{0.2,-0.1,0.2\}$ and $\boldsymbol{\tau}_\phi = \{0.5,0.5,0.5\}$. }
\label{GT_parameters_2}
\end{figure}
\begin{figure}[!t]
\centering
\includegraphics[scale=0.28]{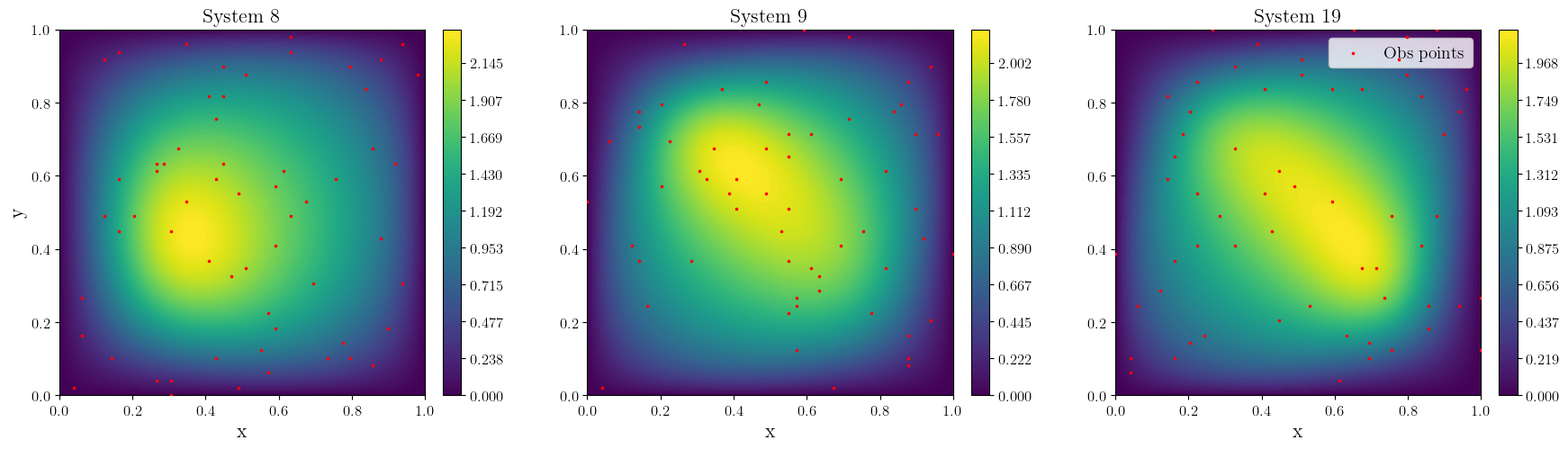}
\caption{ \textbf{Representative ground truth Darcy flow fields generated using fixed-point numerical solver.} The plots display the fields for three selected systems (System 8, 9, and 19) from the dataset of $K=20$ systems. The red markers indicate spatial locations of observation data. }
\label{GT_systems_2}
\end{figure}

While supervised FNO demonstrates superior accuracy in previous metrics, it has the highest computational cost, which stems from the reliance on the expensive numerical solver to generate ground-truth reference trajectories required for the supervised loss calculation during online training. Physics-based FNO improves upon this, achieving runtimes lower than the baseline using numerical solver. In contrast, PINNs emerge as the most efficient architecture, demonstrating a significant speedup compared to the solver. Most notably, PINNs exhibit superior scalability with respect to the number of systems $K$. As observed in Table \ref{tab:result1_efficiency}, while runtimes for the solver and FNO models increase significantly with $K$, the computational cost for PINNs remains approximately invariant. This indicates that PINNs are particularly well-suited for high-dimensional hierarchical inference tasks involving large populations, where they maintain computational feasibility without hugely compromising speed.

\subsection{Nonlinear Darcy Flow}\label{subsec8}

Next, we consider a 2D nonlinear Darcy flow problem. In this setup, the spatially varying permeability field is formulated as the combination of two components: a spatial term defined by three unknown parameters projected onto known basis functions, and an unknown nonlinear closure model that depends on the solution field.

\subsubsection{Problem Formulation}

The two-dimensional nonlinear Darcy flow over $\Omega = [0,1]^2$ takes the form
\begin{equation}
\begin{aligned}
    \label{eq:poisson_problem}
  - \nabla (a(u,\mathbf{x}) \nabla u) &= s(\mathbf{x}), \quad && \mathbf{x} \in \Omega, \\
    u(\mathbf{x}) &= 0, \quad && \mathbf{x} \in \partial \Omega,
\end{aligned}
\end{equation}
where $s(\mathbf{x}) = 2\pi^2 \sin(\pi x)\sin(\pi y)$, $u(\mathbf{x})$ denotes the solution field and $ a(u,\mathbf{x})$ is a nonlinear, spatially varying permeability field, defined as
\begin{align}
    \label{eq:permeability}
    a(u,\mathbf{x}) = \exp(z(\mathbf{x})) \, \boldsymbol{\sigma} (f(u)),
\end{align}
where $\boldsymbol{\sigma}(\cdot)$ is a sigmoid function and the unknown nonlinear closure model is given by $f(u) = u^2 / 2$. The spatial term $z(\mathbf{x})$ is represented using a three-term basis expansion
\begin{align}
    \label{eq:z_x_formulation}
    z(\mathbf{x}) = \sum_{i=1}^{3}z_i\phi_i(\mathbf{x}) ,
\end{align}
% \begin{align}
%     \label{eq:basis_fucntions}
%     \phi_1(\mathbf{x}) &= \sin(2\pi x) \sin(2\pi y) , \\
%     \phi_2(\mathbf{x}) &= \sin(2\pi x) \sin(\pi y) , \\
%     \phi_3(\mathbf{x}) &= \sin(\pi x) \sin(2\pi y) .
% \end{align}
with
\begin{align}
    \label{eq:basis_fucntions}
    \phi_i(\mathbf{x}) &= \sin(m_i\pi x) \sin(n_i\pi y) , 
\end{align}
and $\{ m_1,m_2,m_3\} = \{ 2,2,1\}$ and $\{ n_1,n_2,n_3\}=\{ 2,1,2\}$.

An exponential transform is applied to $z(\mathbf{x})$ to guarantee positivity of the permeability field. The coefficients $\boldsymbol{\theta} = \{z_1, z_2,z_3\}$ are the unknown parameters to be inferred. As in the previous experiment, we impose a hierarchical prior over these parameters with ground-truth population hyperparameters $\boldsymbol{\mu}_\phi=\{ 0.2,-0.1,0.2\}$ and $\boldsymbol{\tau}_\phi = \{ 0.5,0.5,0.5\}$. Generated parameters $\boldsymbol{\theta}$ from the true hyperprior for $K=20$ systems are shown in Figure \ref{GT_parameters_2}.

\subsubsection{Solver} \label{subsecsolver}

To solve the nonlinear Darcy flow problem in \eqref{eq:poisson_problem}, we employ the Finite Element Method (FEM) \citep{ern2004theory}. By deriving the weak formulation and projecting the problem onto a finite-dimensional subspace defined by a triangulation of the domain, we obtain the following nonlinear algebraic system:
\begin{equation}
    \label{eq:nonlinear_matrix_system}
    \mathbf{A}(\mathbf{u}) \mathbf{u} = \mathbf{b},
\end{equation}
where $\mathbf{u}$ is the vector of unknown nodal coefficients, $\mathbf{b}$ is the load vector, and $\mathbf{A}(\mathbf{u})$ is the stiffness matrix. Crucially, due to the dependence of the permeability field $a(u, \mathbf{x})$ on the solution, the stiffness matrix is state-dependent, making the system nonlinear. The detailed derivation of the weak formulation and explicit definitions of the matrix entries are provided in Appendix \ref{secB}.

\begin{figure}[!t]
\centering
\includegraphics[scale=0.28]{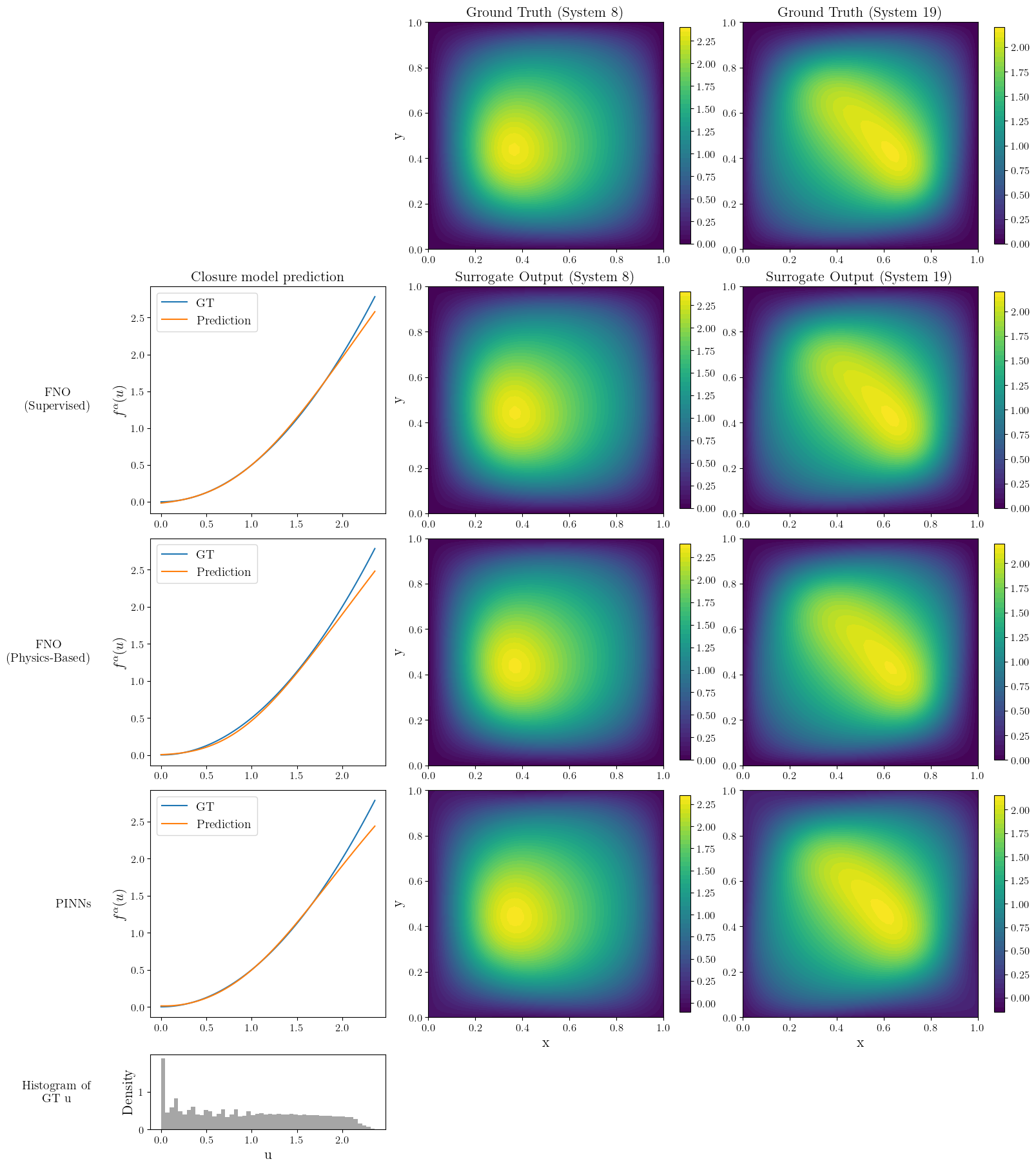}
\caption{ \textbf{Comparison of closure model estimation and surrogate performance across different models with $\mathbf{K=20}$ in Experiment 2.} The rows correspond to GT field generated by fixed-point solver, and results from Supervised FNO, Physics-Based FNO, and PINNs, respectively. \textbf{Left Column:} Comparison between the learned closure model $f^\alpha(u)$ (orange) and the ground truth $f(u)$ (blue). \textbf{Middle $\&$ Right Columns:} Surrogate predictions for two systems (System 8 and System 19). \textbf{Bottom Panel:} Histogram of $u$ in GT dataset. }
\label{alpha_and_beta_2}
\end{figure}
% Since the stiffness matrix $\mathbf{A}$ depends on the unknown vector $\mathbf{u}$, \eqref{eq:nonlinear_matrix_system} represents a nonlinear system. 
To solve this nonlinearity, we employ a fixed-point iteration solver. From an initial guess $\mathbf{u}_0$, the method iteratively solves the following linearized system for $\mathbf{u}_{k+1}$,
\begin{equation}
    \label{eq:fixed_point_iter}
    \mathbf{A}(\mathbf{u}_k) \mathbf{u}_{k+1} = \mathbf{b}.
\end{equation}
$\mathbf{A}(\mathbf{u}_k)$ is constructed using the known solution $\mathbf{u}_k$ from the previous iteration, making the system linear. This process is repeated until the residual falls below a prescribed tolerance.

Three systems from the 20 systems generated using the fixed-point solver are shown in Figure \ref{GT_systems_2} and a fixed-point solver package provided in JAXOpt$\footnotemark \footnotetext{\url{https://jaxopt.github.io/stable/_autosummary/jaxopt.FixedPointIteration.html}}$ is used for data generation and further training.  In addition, 60 observation points are taken from each system with observation noise $\sigma=0.1$. The observation operator $g^{(k)}$ is different for each system in this experiment.

\subsubsection{Surrogate Implementation Details}

In this experiment, the baseline approach using numerical solver as the forward model is omitted due to the high computational cost of the iterative fixed-point solver within the MALA framework. A detailed discussion on computational efficiency is provided in Section \ref{result3}.

% \begin{figure}[!t]
% \centering
% \includegraphics[scale=0.34]{images/result2_3model_mcmc.png}
% \caption{ \textbf{Trace plots illustrating the evolution of a randomly selected MALA chain for $z_1$ of System 9 across three models with $\mathbf{K=20}$ in Experiment 2}. GT parameter value is indicated using green dashed line.}
% \label{3model_mcmc_2}
% \end{figure}
\begin{figure}[!t]
\centering
\includegraphics[scale=0.3]{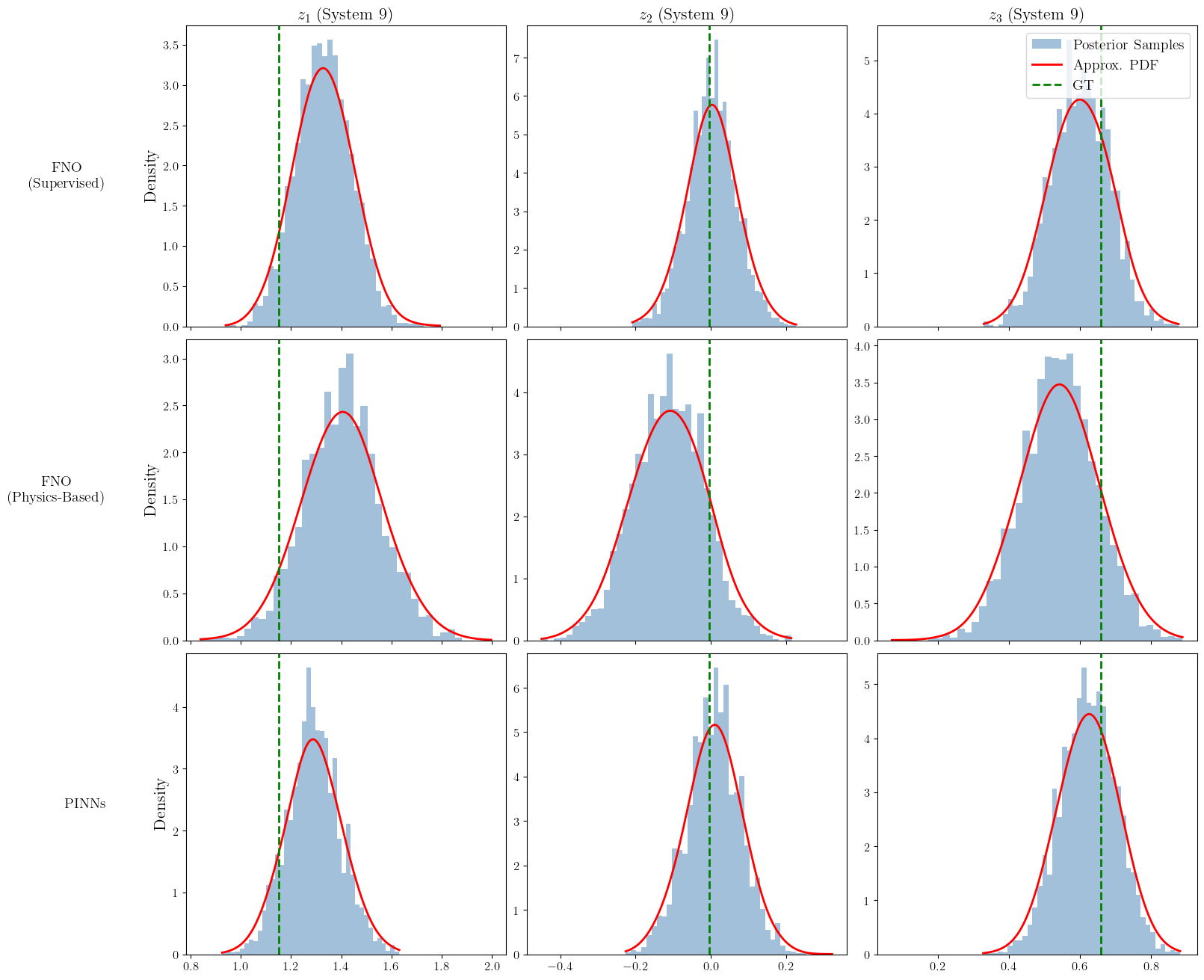}
\caption{ \textbf{Posterior histograms and corresponding approximated posterior density functions for three parameters from System 9 across three models with $\mathbf{K=20}$ in Experiment 2.} Each panel shows the marginal posterior distribution obtained from the final 1,000 samples of one random MALA chain after convergence. }
\label{3model_histogram_2}
\end{figure}
\begin{table}[h]
\centering
\caption{\textbf{Quantitative comparison of three models across varying numbers of systems ($\mathbf{K \in \{10, 20,30\}}$) in Experiment 2.} The table evaluates \textbf{Parameter Inference}, \textbf{Closure MSE} and \textbf{Surrogate MSE} with metrics the same as in Table \ref{tab:result1_models_performance}. Closure MSE is evaluated over range $u \in [0,2.5]$.}
\label{tab:result3_models_performance}
\renewcommand{\arraystretch}{1.3}
\resizebox{0.6\linewidth}{!}{%
\begin{tabular}{ccccccc}
\toprule
\textbf{Model} 
& \makecell{\textbf{No. of}\\\textbf{Systems}}
& \multicolumn{2}{c}{\textbf{Parameter Inference}} 
& \textbf{Closure MSE} 
& \textbf{Surrogate MSE} 
\\
\cmidrule(lr){3-4}
& & \textbf{Mean MSE} & \textbf{Coverage} & & \\ 
\midrule

\multirow{3}{*}{\makecell{\textbf{FNO}\\\textbf{(Supervised)}}}
 & $K = 10$ & $3.39 \times 10^{-2}$ & $93.3 \%$ & $5.01 \times 10^{-3}$ & $8.92 \times 10^{-4}$ \\
 & $K = 20$ & $1.43 \times 10^{-2}$ & $93.3 \%$ & $3.95 \times 10^{-3}$ & $3.48 \times 10^{-5}$ \\
 & $K = 30$ & $2.37 \times 10^{-2}$ & $90.0 \%$ & $3.28 \times 10^{-3}$ & $3.57 \times 10^{-5}$ \\
\midrule

\multirow{3}{*}{\makecell{\textbf{FNO}\\\textbf{(Physics-based)}}}
 & $K = 10$ & $1.17 \times 10^{-1}$ & $56.7 \%$ & $4.28 \times 10^{-2}$ & $7.20 \times 10^{-3}$\\
 & $K = 20$ & $2.47 \times 10^{-2}$ & $91.7 \%$ & $1.49 \times 10^{-2}$ & $6.35 \times 10^{-5}$\\
 & $K = 30$ & $3.00 \times 10^{-2}$ & $88.9 \%$ & $9.27 \times 10^{-3}$ & $8.52 \times 10 ^{-4}$ \\
\midrule

\multirow{3}{*}{\textbf{PINNs}}
 & $K = 10$ & $4.89 \times 10^{-2}$ & $83.3 \%$ & $2.62 \times 10^{-1}$ & $1.09 \times 10^{-3}$ \\
 & $K = 20$ & $1.53 \times 10^{-2}$ & $93.3 \%$ & $8.31 \times 10^{-3}$ & $1.56 \times 10^{-4}$ \\
 & $K = 30$ & $2.56 \times 10^{-2}$ & $90.0 \%$ & $1.11 \times 10^{-2}$ & $9.33 \times 10^{-4}$ \\
\bottomrule

\end{tabular}%
}
\end{table}

We utilize the same three surrogate model architectures as in the first experiment, with the exception that the inputs are now spatial coordinates $x$ and $y$, rather than time $t$, to suit this stationary PDE problem. For the online training of surrogate models, the loss function of supervised FNO is the discrepancy between surrogate predictions and reference solutions obtained from the fixed-point solver. For physics-based FNO, we employ the weak-form residual loss as defined in \eqref{eq:fno_physics_loss}, where
\begin{equation}
    \label{eq:weak_loss}
    \mathcal{R} = \mathbf{A}(\mathbf{u}) \mathbf{u} - \mathbf{b},
\end{equation}
consistent with the discretized system in \eqref{eq:nonlinear_matrix_system}. Finally, the PINN loss is computed directly from the strong form equation \eqref{eq:poisson_problem} using automatic differentiation. More implementation details could be found in Appendix \ref{secA}.

\subsubsection{Results} \label{result3}

As in the first experiment, our evaluation follows the same three-part structure: a performance comparison of the four models, a contrast between the hierarchical and non-hierarchical Bayesian settings, and an analysis of computational efficiency.

\vspace{1ex}
\noindent\textbf{Performance Comparison on Inverse and Forward Problems}

To evaluate the performance of closure and surrogate models, Figure \ref{alpha_and_beta_2} visualizes estimated closure models and surrogate outputs for all three architectures at $K=20$. A corresponding quantitative analysis across varying number of systems is detailed in Table \ref{tab:result3_models_performance}. Supervised FNO consistently outperforms physics-informed models, notably, its surrogate MSE is generally an order of magnitude lower than that of the other methods, accompanied by a highly accurate reconstruction of the closure. Conversely, both physics-based FNO and PINNs exhibit higher MSE, suggesting that relying solely on physics-informed residuals for training surrogates in this 2D setting is less stable than direct supervision. This degradation is most severe at $K=10$, where the limited number of observed systems likely fails to provide sufficient constraints for physics-based optimization to converge. Furthermore, unlike FNO architectures which naturally encode boundary structures, PINNs impose boundary conditions as soft constraints as mentioned in \eqref{eq:PINNs_loss}. As observed in Figure \ref{alpha_and_beta_2}, this leads to errors at domain boundaries of surrogate outputs. 

\begin{figure}[!t]
\centering
\includegraphics[scale=0.36]{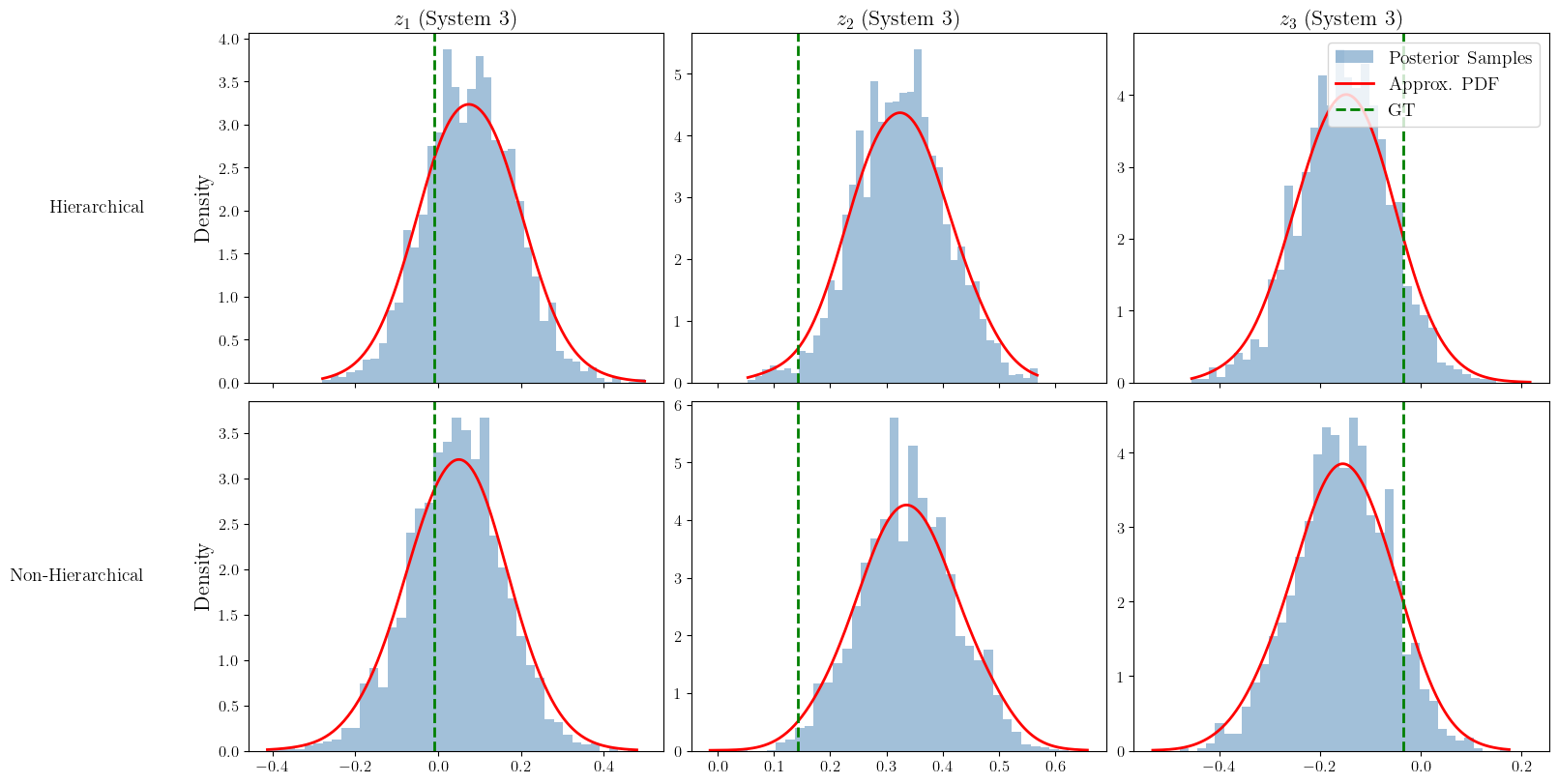}
\caption{ \textbf{Comparison of posterior parameter distributions between Hierarchical and Non-Hierarchical Bayesian frameworks using a PINN-based surrogate with $\mathbf{K=20}$ in Experiment 2. } The figure compares the uncertainty quantification performance for three parameters from System 3 for both hierarchical and non-hierarchical setups. }
\label{hierarchical_histogram_2}
\end{figure}

\begin{table}[t!]
\centering
\caption{\textbf{Comparison of hierarchical and non-hierarchical approaches trained using PINNs as surrogate model in Experiment 2.} Reported metrics are the same as in Table \ref{tab:result1_models_performance}.}
\renewcommand{\arraystretch}{1.3}
\resizebox{0.6\linewidth}{!}{%
\begin{tabular}{ccccccc}
\toprule
\label{tab:hierarchical_comparison_3}
\textbf{Method} 
& \makecell{\textbf{No. of}\\\textbf{Systems}}
& \multicolumn{2}{c}{\textbf{Parameter Inference}}
& \textbf{Closure MSE}
& \textbf{Surrogate MSE}
\\
\cmidrule(lr){3-4}
& & \textbf{Mean MSE} & \textbf{Coverage} & & \\
\midrule

\multirow{3}{*}{\textbf{Hierarchical}}
 & $K = 10$ & $4.89 \times 10^{-2}$ & $83.3 \%$ & $2.62 \times 10^{-1}$ & $1.09 \times 10^{-3}$ \\
 & $K = 20$ & $1.53 \times 10^{-2}$ & $93.3 \%$ & $8.31 \times 10^{-3}$ & $1.56 \times 10^{-4}$ \\
 & $K = 30$ & $2.56 \times 10^{-2}$ & $90.0 \%$ & $1.11 \times 10^{-2}$ & $9.33 \times 10^{-4}$ \\
\midrule

\multirow{3}{*}{\makecell{\textbf{Non-}\\\textbf{Hierarchical}}}
 & $K = 10$ & $6.21 \times 10^{-2}$ & $80.0 \%$ & $5.33 \times 10^{-1}$ & $1.69 \times 10^{-3}$\\
 & $K = 20$ & $1.46 \times 10^{-2}$ & $93.3 \%$ & $9.62 \times 10^{-3}$ & $1.79 \times 10^{-4}$\\
 & $K = 30$ & $3.73 \times 10^{-2}$ & $82.2 \%$ & $1.52 \times 10^{-2}$ & $1.69 \times 10^{-3}$ \\
% \midrule

\bottomrule
\end{tabular}%
}
\end{table}
% \begin{figure}[!t]
% \centering
% \includegraphics[scale=0.4]{images/result2_hyperprior.png}
% \caption{ \textbf{MALA chains of posterior samples for hyperparameters inferred using the PINN-based hierarchical Bayesian framework with $\mathbf{K=20}$ in Experiment 2.} The panels display the MALA chains for $\boldsymbol{\mu}_\phi$ and $\boldsymbol{\tau}_\phi$. Green dashed lines represent GT values. }
% \label{hypeprior_2}
% \end{figure}

Regarding parameter inference, 
% Figure \ref{3model_mcmc_2} illustrates a representative MALA trace plot for parameter $z_1$ in System 9, while 
Figure \ref{3model_histogram_2} displays histograms of final 1,000 posterior samples for all three parameters in System 9. Again, quantitative analysis of accuracy and coverage of parameter inference is given in Table \ref{tab:result3_models_performance}. The results clearly establish supervised FNO as the most accurate and robust method. Its superiority is particularly critical in the limited data regime ($K=10$). In this setting, purely physics-informed approaches fail to reliably identify parameters, exhibiting high mean MSE and poor coverage probabilities ($56.7\%$ and $83.3\%$). In contrast, supervised FNO maintains high accuracy and reliable uncertainty quantification in this scenario, achieving a low MSE and a coverage of $93.3\%$.

\vspace{1ex}
\noindent\textbf{Hierarchical and Non-Hierarchical Bayes}

Hierarchical and non-hierarchical Bayesian formulations are compared as well. Again, PINNs are used as the surrogate architecture. Figure \ref{hierarchical_histogram_2} visualizes posterior distributions of three parameters in System 3, while Table \ref{tab:hierarchical_comparison_3} details quantitative metrics.

Visually, Figure \ref{hierarchical_histogram_2} indicates that the non-hierarchical model results in wider posterior distributions, reflecting higher uncertainty. In contrast, the hierarchical framework shows a higher degree of confidence in parameter estimates. By conditioning individual parameters on a shared population distribution, the model effectively constrains the variance of specific systems using group-level information. Quantitatively, in Table \ref{tab:hierarchical_comparison_3}, the hierarchical setting consistently outperforms the non-hierarchical formulation across different system counts $K$, in both inverse inference and surrogate training.

\begin{table}[!t]
\centering
\caption{\textbf{Computational efficiency comparison across considered models for varying number of systems ($\mathbf{K \in \{10, 20, 30\}}$) in Experiment 2.}}
\resizebox{0.82\linewidth}{!}{%
\begin{tabular*}{\textwidth}{@{\extracolsep{\fill}}cccccc}
\toprule
\textbf{Samples} & \textbf{Solver} & \makecell{\textbf{FNO}\\\textbf{(Supervised)}} & \makecell{\textbf{FNO}\\\textbf{(Physics-based)}} & \textbf{PINNs} & \makecell{\textbf{PINNs}\\\textbf{(Non-Hierarchical)}} \\
\midrule
$K=10$ & $4.158s$ & $3.121s$ & $0.686s$ & $0.132s$ & $0.134s$ \\
$K=20$ & $10.374s$ & $3.837s$ & $1.119s$ & $0.222s$ & $0.223s$ \\
$K=30$ & $17.962s$& $4.962s$ & $1.934s$ & $0.378s$ & $0.381s$ \\
\bottomrule
\label{tab:result2_efficiency}
\end{tabular*}%
}
\end{table}

Inferred hyperprior information from the hierarchical setting when $K=20$ is $\overline{\boldsymbol{\mu}}_\phi = \{ 0.37, -0.09, -0.01 \}$ and $\overline{\boldsymbol{\tau}}_\phi = \{ 0.35, 0.29, 0.28\}$.
%, with corresponding MALA trajectories shown in Figure \ref{hypeprior_2}. 
Ground truth values are $\boldsymbol{\mu}_\phi = \{ 0.2, -0.1,0.2\}$ and $\boldsymbol{\mu}_\phi = \{ 0.5, 0.5, 0.5\}$. A noticeable deviation is observed in hyperprior variances $\boldsymbol{\tau}_\phi$. This discrepancy is statistically attributable to the limited sample size ($K=20$), because accurate estimation of variance inherently requires a larger number of observed systems.

\vspace{1ex}
\noindent\textbf{Efficiency of different algorithms}

Finally, we assess the computational efficiency of proposed frameworks, with results summarized in Table \ref{tab:result2_efficiency}. Consistent with the methodology in Experiment 1, the reported runtime measures the average duration of one complete training epoch, including both the inverse inference step (ensemble MALA and closure update) and the surrogate training step. Notably, the baseline approach using traditional numerical solver as forward evaluation exhibits significant computational costs in this nonlinear PDE problem (e.g., $17.96$s per epoch at $K=30$).

In contrast, all surrogate-based approaches demonstrate substantial speedups. PINNs again emerge as the most computationally efficient architecture, maintaining low runtimes with excellent scalability. However, considering the suboptimal performance observed in purely physics-based methods in this experiment, supervised FNO provides the most balanced solution. While computationally more demanding than PINNs, it still offers a significant reduction in runtime compared to the numerical solver while delivering higher accuracy required for robust parameter and closure discovery.

\subsection{Generalized Burgers' Equation}

Finally, the last experiment evaluates a generalized Burgers' equation. In this setup, the unknown parameters to be inferred are viscosity and the initial condition amplitude, while the convective term is treated as the unknown closure.

\begin{figure}[!t]
\centering
\includegraphics[scale=0.4]{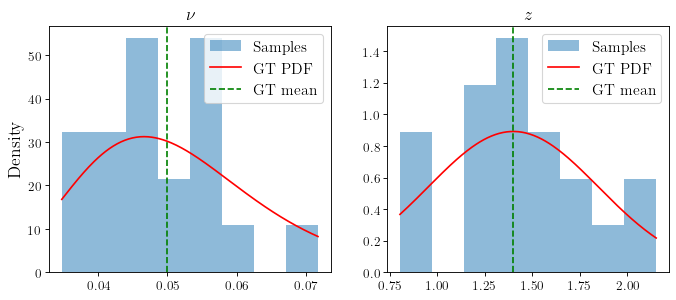}
\caption{ \textbf{Distribution of ground-truth physical parameters for 20 Burgers' systems in Experiment 3.} Each system’s parameter vector $\boldsymbol{\theta} = \{ \log(\nu), z\}$ is sampled from the hierarchical prior with hyperparameters $\boldsymbol{\mu}_\phi = \{ \log(0.05), 1.4\}$ and $\boldsymbol{\tau}_\phi = \{ 0.07, 0.2\}$.  }
\label{parameters_3}
\end{figure}
\begin{figure}[!t]
\centering
\includegraphics[scale=0.26]{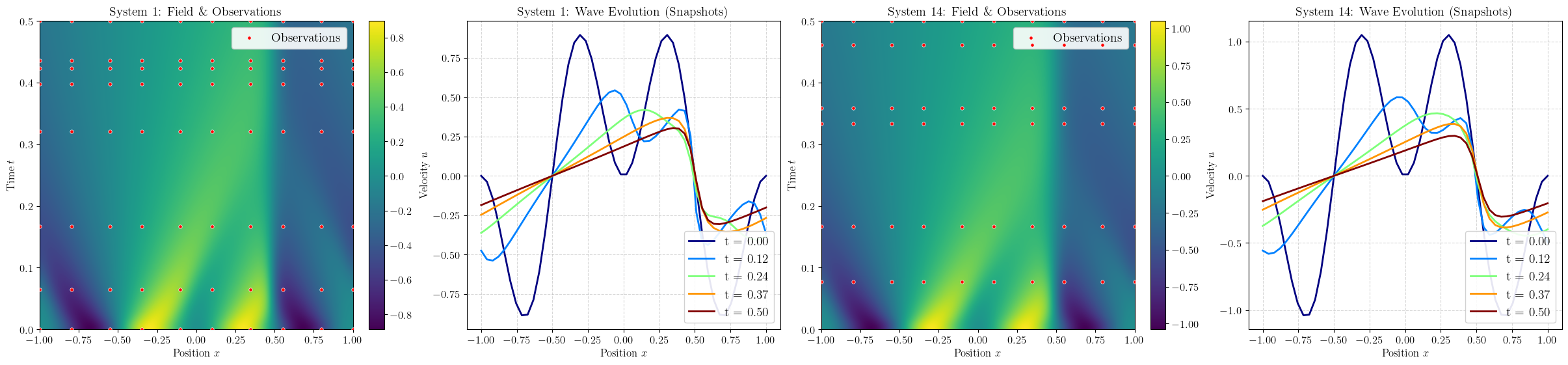}
\caption{\textbf{Representative ground-truth Burgers' fields generated using numerical solver.} The plots display the fields for two selected systems (System 1 and 14) from the dataset of $K=20$ systems. The red markers indicate spatial locations of observation data.}
\label{fields_3}
\end{figure}

\subsubsection{Problem Formulation}

A generalized Burgers' equation has the form
\begin{equation}
    \begin{aligned}
    \label{eq:burgers}
    \frac{\partial u}{\partial t} + f(u) \frac{\partial u}{\partial x} &= \nu \frac{\partial ^2 u}{\partial x^2}, \quad x\in [-1,1], \ t \in [0,0.5], \\
    u(x,0) &= z \sin (2 \pi x) \sin (\pi x), \\
    u(-1,t) &= u(1,t),
    \end{aligned}
\end{equation}
where periodic boundary conditions are applied. $u(x,t)$ represents the velocity solution field. $f(u)$ is a nonlinear function of $u$ with $f(u) = 7 (\boldsymbol{\sigma}(3 u) - 0.5)$, where $\boldsymbol{\sigma}(\cdot)$ is a sigmoid function. In this case, both the viscosity $\nu$ and $z$ in the initial condition are unknown parameters. Because $\nu$ has to be positive, parameters to be inferred are set to $\boldsymbol{\theta} = \{ \log(\nu), z\}$. The ground-truth hyperprior of the population is $\boldsymbol{\mu}_\phi = \{ \log(0.05), 1.4\}$ and $\boldsymbol{\tau}_\phi = \{ 0.07, 0.2\}$. Parameters of generated $K=20$ systems are shown in Figure \ref{parameters_3}.

\subsubsection{Solver}

\begin{figure}[!t]
\centering
\includegraphics[scale=0.36]{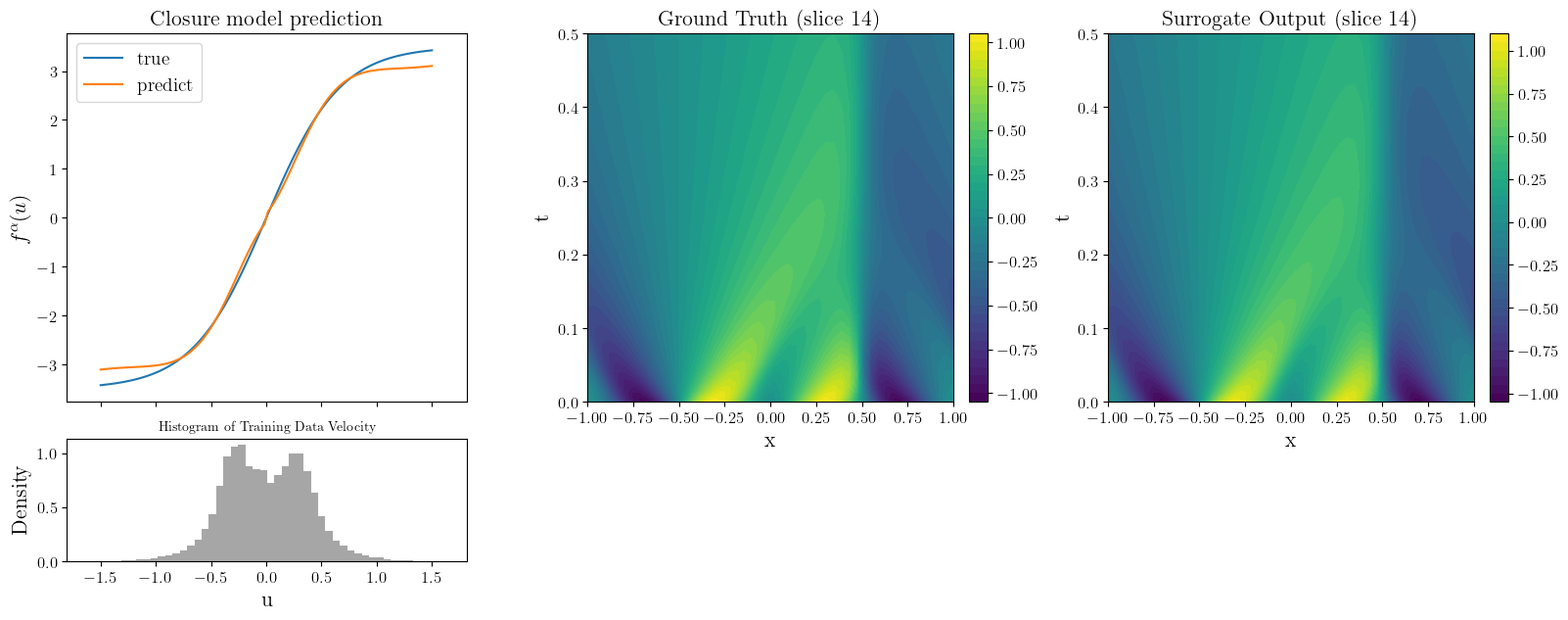}
\caption{ \textbf{Closure model estimation and surrogate performance of supervised FNO with $\mathbf{K=20}$ in Experiment 3.} \textbf{Left:} Closure model estimation with histograms of ground-truth velocity data provided. \textbf{Middle $\&$ Right: } Ground truth and surrogate prediction of solution field of System 14.}
\label{alpha_and_beta_3}
\end{figure}

\begin{figure}[!t]
\centering
\includegraphics[scale=0.32]{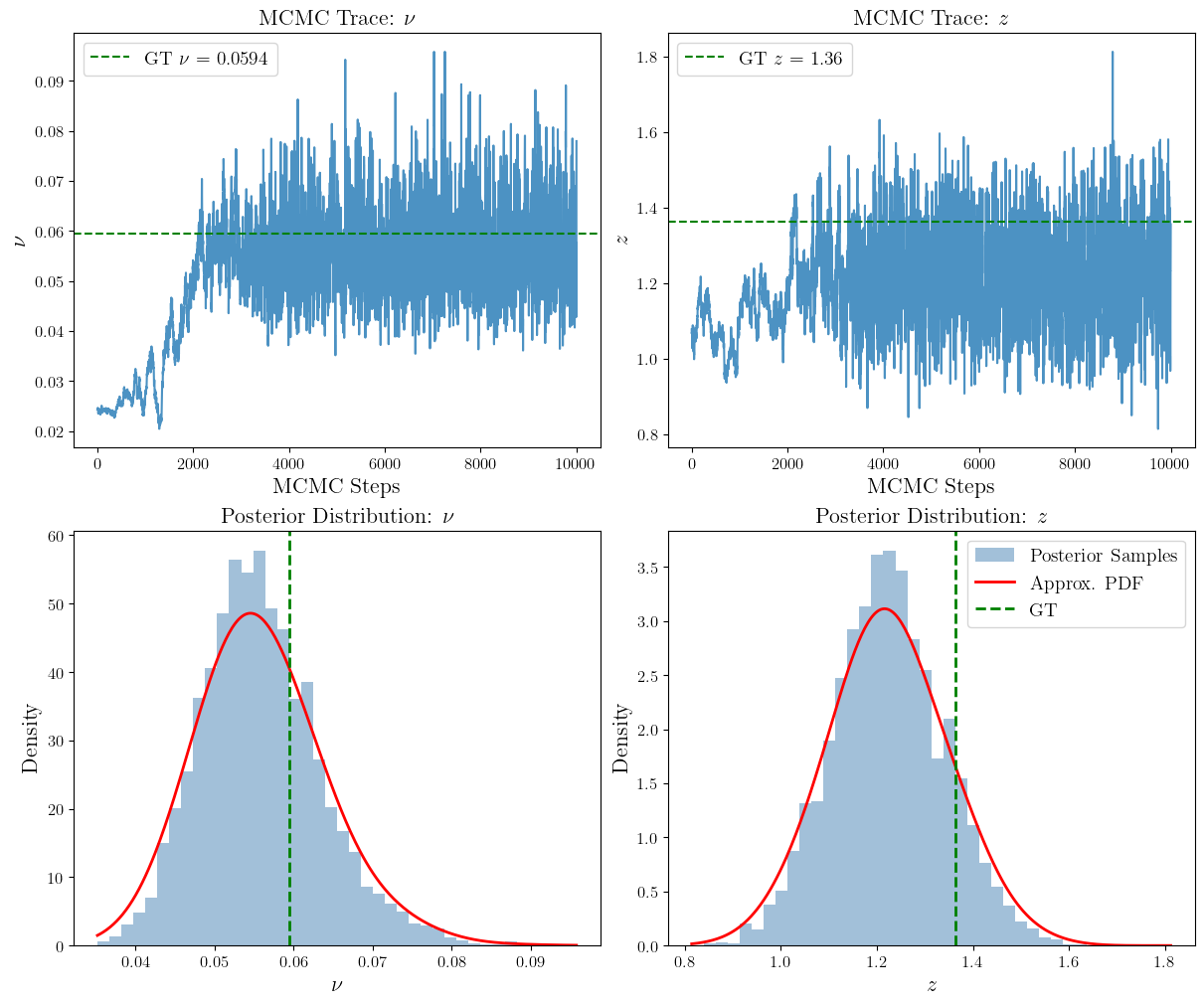}
\caption{ \textbf{MALA traces and posterior histograms of parameters from System 14 in Experiment 3. } \textbf{Top row:} MALA traces of parameters $\nu$ and $z$. \textbf{Bottom row:} posterior histograms of last 1,000 samples of MALA traces and their corresponding PDF approximations.  }
\label{MCMC_3}
\end{figure}

To numerically solve \eqref{eq:burgers}, we employ an explicit fourth-order Runge–Kutta (RK4) method for time integration with a time step of $\Delta t = 0.00125s$ and a spatial resolution of $\Delta x = 0.04$. To reduce storage overhead, the solution is subsampled by recording data every 10 temporal steps, yielding a total of 40 discrete snapshots across the time domain. Notably, the restrictive numerical setup required to simulate such time-dependent systems highlights a primary advantage of utilizing a surrogate model for the forward pass. Explicit numerical solvers require exceptionally small timesteps to maintain numerical stability, necessitating a vast number of sequential iterations. During the inverse inference phase, backpropagating through these extensive solver iterations to compute gradients becomes computationally prohibitive. In contrast, the surrogate model circumvents the need for dense iterative time-stepping, drastically accelerating the gradient computation for the inverse problem.

In this experiment, system-specific observation operators $g^{(k)}$ are employed as well. Measurements are recorded at fixed spatial locations but at randomly sampled time steps, with the total number of temporal observations varying between 5 and 10 per system. Figure \ref{fields_3} shows the solution fields, wave evolutions, and corresponding observation locations for Systems 1 and 14. All measurements are subject to an observation noise level of $\sigma = 0.2$.

\subsubsection{Results}

Only FNO trained in a supervised way is implemented in this case with $K=20$ systems, and more implementation details are provided in Appendix \ref{secA}. 

The estimated closure model and surrogate outputs are provided in Figure \ref{alpha_and_beta_3}. The MSE for closure model estimation evaluated at $u \in [-1.5,1.5]$ is $2.90 \times 10 ^{-2}$ and MSE of surrogate outputs across all systems is $1.49 \times 10^{-4}$. Results for parameter inference of one selected System 14 are given in Figure \ref{MCMC_3}, showing MALA traces and posterior histograms of both unknown parameters. Across $K=20$ systems, MSE between estimated posterior means and ground-truth parameters is $1.64 \times 10^{-2}$ and the coverage of estimated posterior distributions within 2 standard deviation is $95 \%$. The mean of posterior samples of the hyperprior is $\overline{\boldsymbol{\mu}}_\phi = \{ \log(0.051), 1.42 \}$ and $\overline{\boldsymbol{\tau}}_\phi = \{ 0.029, 0.176\}$, compared with $\boldsymbol{\mu}_\phi = \{ \log(0.05), 1.4\}$ and $\boldsymbol{\tau}_\phi = \{ 0.07, 0.2\}$.

\section{Conclusion and Future Work}\label{sec6}

In this work, we present a comprehensive bilevel optimization framework for simultaneous resolution of forward and inverse problems in differential equations. Our approach specifically targets scenarios governed by partially known physical laws, where both unknown scalar parameters and nonlinear closures are inferred from sparse observations. Our approach decouples the inference task, combining a hierarchical Bayesian framework for probabilistic parameter estimation with deterministic maximum likelihood closure learning for unknown dynamics. By employing ensemble MALA with covariance preconditioning, we efficiently sample from the posterior, utilizing the samples generated across the ensemble chains to facilitate accurate closure learning.

To mitigate the computational cost of repeated forward evaluations, we integrate a jointly trained surrogate model within the optimization loop. Our experimental results across ODE and PDE systems validate the effectiveness of the proposed framework and highlight critical trade-offs between surrogate architectures. In lower-dimensional ODE settings, PINNs offer a competitive balance of accuracy and efficiency. For complex problems like nonlinear Darcy flow and generalized Burgers', surrogate modeling becomes essential for computational feasibility. In this regime, supervised FNO demonstrates superior robustness and stability compared to physics-based alternatives, particularly in limited data scenarios.

In summary, this work establishes a flexible pathway for parameter inference and closure learning from sparse observations by effectively coupling probabilistic inference, neural closure learning, and surrogate model training. A direction for future research is incorporating techniques like Kalman filtering to facilitate online, joint state and parameter inference in dynamical systems.

\begin{appendices}

\section{Experimental Details}\label{secA}

All experiments are implemented in JAX and executed on an NVIDIA GeForce RTX 3090 GPU. In this section, we provide detailed hyperparameter settings and architectural specifications for the proposed framework. Implementation details for neural network architectures, MALA sampler and optimization schemes of all experiments are listed below.

We utilize three distinct network architectures: FNO and PINNs for the surrogate model $F^\beta$ (sharing the same architecture for both supervised and physics-based training), and an MLP for the closure model $f^\alpha$. All models are trained using the Adam optimizer \citep{kingma2014adam}. Tables \ref{tab:NN_details_1} and \ref{tab:training_params_1} outline the network structures and optimization hyperparameters for Experiment 1, while Tables \ref{tab:NN_details_2} and \ref{tab:training_params_2} present the corresponding settings for Experiment 2, and Tables \ref{tab:NN_details_3} and \ref{tab:training_params_3} are for Experiment 3.

% \begin{table}[H]
%     \centering
\vspace{1em}
\noindent
    \begin{minipage}[c]{0.52\linewidth}
        % \centering
        \captionof{table}{Network architectures for surrogate and closure models in Experiment 1.}
        \label{tab:NN_details_1}
        \resizebox{\linewidth}{!}{%
        \begin{tabular}{c c c}
            \toprule
            \textbf{Model} & \textbf{Hyperparameter} & \textbf{Specification} \\
            \midrule
            \multirow{6}{*}{\textbf{FNO}} & Lifted Dimension (Width) & 32 \\
                                         & Number of Modes & 16 \\
                                         & Number of FNO Layers & 4 \\
                                         & Input Dimension & 4 \\
                                         & Output Dimension & 2 \\
                                         & Activation Function & SiLU \\
            \midrule
            \multirow{5}{*}{\textbf{PINNs}} & Hidden Layers & 5 \\
                           & Hidden Neurons per Layer & 64 \\
                           & Input Dimension & 4 \\
                           & Output Dimension & 1 \\
                           & Activation Function & Tanh \\
            \midrule
            \multirow{5}{*}{\textbf{MLP (Closure Model)}} & Hidden Layers & 4 \\
                                         & Hidden Neurons per Layer & 64 \\
                                         & Input Dimension & 1 \\
                                         & Output Dimension & 1 \\
                                         & Activation Function & SiLU \\
            \bottomrule
        \end{tabular}%
        }
    \end{minipage}\hfill 
    \begin{minipage}[c]{0.44\linewidth}
        % \centering
        \captionof{table}{Hyperparameter settings for optimization and training in Experiment 1.}
        \label{tab:training_params_1}
        \resizebox{\linewidth}{!}{%
        \begin{tabular}{c c c}
            \toprule
            \textbf{Hyperparameter} & \textbf{FNO} & \textbf{PINNs} \\
            \midrule
            Number of MALA Chains ($M$) & 200 & 200 \\
            MALA Step Size ($\gamma$) & 0.010 & 0.018 \\
            Lower-Level Iterations ($N$) & 20 & 20 \\
            Closure Learning Rate ($\eta_\alpha$) & $5 \times 10^{-4}$ & $1 \times 10^{-3}$  \\
            Surrogate Learning Rate ($\eta_\beta$) & $5 \times 10^{-4}$ & $1 \times 10^{-3}$ \\
            \bottomrule
        \end{tabular}% 
        }
    \end{minipage}
% \end{table}

% \begin{table}[H]
%     \centering
\vspace{1em}
\noindent
    \begin{minipage}[c]{0.52\linewidth}
        % \centering
        \captionof{table}{Network architectures for surrogate and closure models in Experiment 2.}
        \label{tab:NN_details_2}
        \resizebox{\linewidth}{!}{%
        \begin{tabular}{c c c}
            \toprule
            \textbf{Model} & \textbf{Hyperparameter} & \textbf{Specification} \\
            \midrule
            \multirow{6}{*}{\textbf{FNO}} & Lifted Dimension (Width) & 32 \\
                                         & Number of Modes & 16 \\
                                         & Number of FNO Layers & 4 \\
                                         & Input Dimension & 5 \\
                                         & Output Dimension & 1 \\
                                         & Activation Function & SiLU \\
            \midrule
            \multirow{5}{*}{\textbf{PINNs}} & Hidden Layers & 5 \\
                           & Hidden Neurons per Layer & 64 \\
                           & Input Dimension & 5 \\
                           & Output Dimension & 1 \\
                           & Activation Function & Tanh \\
            \midrule
            \multirow{5}{*}{\textbf{MLP (Closure Model)}} & Hidden Layers & 4 \\
                                         & Hidden Neurons per Layer & 64 \\
                                         & Input Dimension & 1 \\
                                         & Output Dimension & 1 \\
                                         & Activation Function & SiLU \\
            \bottomrule
        \end{tabular}%
        }
    \end{minipage}\hfill 
    \begin{minipage}[c]{0.44\linewidth}
        % \centering
        \captionof{table}{Hyperparameter settings for optimization and training in Experiment 2.}
        \label{tab:training_params_2}
        \resizebox{\linewidth}{!}{%
        \begin{tabular}{c c c}
            \toprule
            \textbf{Hyperparameter} & \textbf{FNO} & \textbf{PINNs} \\
            \midrule
            Number of MALA Chains ($M$) & 160 & 160 \\
            MALA Step Size ($\gamma$) & 0.018 & 0.020 \\
            Lower-Level Iterations ($N$) & 18 & 20 \\
            Surrogate Learning Rate ($\eta_\beta$) & $5 \times 10^{-4}$ & $5 \times 10^{-4}$ \\
            Closure Learning Rate ($\eta_\alpha$) & $5 \times 10^{-4}$ & $5 \times 10^{-4}$  \\
            \bottomrule
        \end{tabular}% 
        }
    \end{minipage}

\vspace{1em}
\noindent
% \begin{table}[H]
%     \centering
    \begin{minipage}[c]{0.52\linewidth}
        \centering
        \captionof{table}{Network architectures for surrogate and closure models in Experiment 3.}
        \label{tab:NN_details_3}
        \resizebox{\linewidth}{!}{%
        \begin{tabular}{c c c}
            \toprule
            \textbf{Model} & \textbf{Hyperparameter} & \textbf{Specification} \\
            \midrule
            \multirow{6}{*}{\textbf{FNO}} & Lifted Dimension (Width) & 32 \\
                                         & Number of Modes & 16 \\
                                         & Number of FNO Layers & 4 \\
                                         & Input Dimension & 4 \\
                                         & Output Dimension & 1 \\
                                         & Activation Function & SiLU \\
            \midrule
            \multirow{5}{*}{\textbf{MLP (Closure Model)}} & Hidden Layers & 2 \\
                                         & Hidden Neurons per Layer & 64 \\
                                         & Input Dimension & 1 \\
                                         & Output Dimension & 1 \\
                                         & Activation Function & SiLU \\
            \bottomrule
        \end{tabular}%
        }
    \end{minipage}\hfill 
    \begin{minipage}[c]{0.44\linewidth}
        % \centering
        \captionof{table}{Hyperparameter settings for optimization and training in Experiment 3.}
        \label{tab:training_params_3}
        \resizebox{0.81\linewidth}{!}{%
        \begin{tabular}{c c}
            \toprule
            \textbf{Hyperparameter} & \textbf{FNO}  \\
            \midrule
            Number of MALA Chains ($M$) & 120  \\
            MALA Step Size ($\gamma$) & 0.018  \\
            Lower-Level Iterations ($N$) & 18  \\
            Surrogate Learning Rate ($\eta_\beta$) & $5 \times 10^{-4}$ \\
            Closure Learning Rate ($\eta_\alpha$) & $5 \times 10^{-4}$  \\
            \bottomrule
        \end{tabular}%
        }
    \end{minipage}

\section{Weak Formulation and FEM Discretization}\label{secB}

In this section, we provide the mathematical details regarding the weak formulation and discretization of the nonlinear Darcy flow problem in Section \ref{subsecsolver}.

We first derive the weak formulation of the PDE in \eqref{eq:poisson_problem}. We define the trial and test function space as $V = H^1_0(\Omega)$, assuming homogeneous Dirichlet boundary conditions. Multiplying the governing equation by an arbitrary test function $v \in V$ and integrating over the domain $\Omega$ yields
\begin{equation}
    \label{eq:weak_form_1}
    \int_{\Omega} -\nabla \cdot (a(u, \mathbf{x})\nabla u) v \, d\mathbf{x} = \int_{\Omega} s(\mathbf{x}) v \, d\mathbf{x}.
\end{equation}

Applying integration by parts to the term on the left-hand side, we obtain
\begin{equation}
    \label{eq:weak_form_2}
    \int_{\Omega} a(u, \mathbf{x}) \nabla u \cdot \nabla v \, d\mathbf{x} - \int_{\partial \Omega} a(u, \mathbf{x}) (\nabla u \cdot \mathbf{n}) v \, ds = \int_{\Omega} s(\mathbf{x}) v \, d\mathbf{x}.
\end{equation}
Since $v \in H^1_0(\Omega)$ vanishes on the boundary $\partial \Omega$, the boundary integral becomes zero. Thus, the weak formulation is stated as follows: find $u \in V$ such that
\begin{equation} 
    \label{eq:weak_form_3}
    \int_{\Omega} a(u, \mathbf{x}) \nabla u \cdot \nabla v \  d\mathbf{x} = \int_{\Omega} s(\mathbf{x}) v \ d\mathbf{x}, \quad \forall v \in V.
\end{equation}
To approximate the solution numerically, we utilize FEM \citep{ern2004theory} by restricting the problem to a finite-dimensional subspace $V_h \subset V$. We introduce a triangulation of the domain $\Omega$ and define a set of basis functions $\{\psi_j\}_{j=1}^{Q}$, where $Q$ denotes the total number of degrees of freedom. The discrete solution $u_h \in V_h$ is expressed as a linear expansion of these basis functions,
\begin{equation}
    \label{eq:fem_expansion}
    u_h(\mathbf{x}) = \sum_{j=1}^{Q} u_j \psi_j(\mathbf{x}),
\end{equation}
and $\mathbf{u} = [u_1, u_2, \dots, u_Q]^T$ represents the vector of unknown nodal coefficients. Substituting this expansion into the weak formulation \eqref{eq:weak_form_3} and applying the Galerkin method (i.e., $v = \psi_i$ for $i=1, \dots, Q$) yields the following system of nonlinear equations,
\begin{equation}
    \label{eq:weak_systems}
    \sum_{j=1}^{Q} u_j \left( \int_{\Omega} a(u_h, \mathbf{x}) \nabla \psi_j \cdot \nabla \psi_i \, d\mathbf{x} \right) = \int_{\Omega} s(\mathbf{x}) \psi_i \, d\mathbf{x}, \quad \text{for } i=1, \dots, Q.
\end{equation}
This system can be written in compact matrix notation as
\begin{equation}
    \label{eq:nonlinear_matrix_system_2}
    \mathbf{A}(\mathbf{u}) \mathbf{u} = \mathbf{b},
\end{equation}
where $\mathbf{b} \in \mathbb{R}^N$ is the load vector with entries $b_i = \int_{\Omega} s(\mathbf{x}) \psi_i \, d\mathbf{x}$, and $\mathbf{A}(\mathbf{u}) \in \mathbb{R}^{N \times N}$ is the stiffness matrix with entries depending on the solution $\mathbf{u}$,
\begin{equation}
    \label{eq:matrix_A}
    \mathbf{A}_{ij}(\mathbf{u}) = \int_{\Omega} a(u_h, \mathbf{x}) \nabla \psi_j \cdot \nabla \psi_i \, d\mathbf{x}.
\end{equation}

\end{appendices}

\section*{Acknowledgements}
We are thankful to Tim J. Rogers for insightful discussions around system identification. AV is supported through the EPSRC ROSEHIPS grant [EP/W005816/1]. AGD is supported through the EPSRC UbOHT Programme grant [EP/X037770/1]. MG is supported by a Royal Academy of Engineering Research Chair and EPSRC grants [EP/X037770/1, EP/Y028805/1, EP/W005816/1, EP/V056441/1 and EP/V056522/1].

\section*{Code Availability}
The source code used during the current study is available in the GitHub repository, \url{https://github.com/ypzpy/Hierarchical-Inference-and-Closure-Learning.git}.

\bibliographystyle{elsarticle-num}
% \bibliographystyle{unsrtnat}

\bibliography{arxiv/reference}

\end{document}